\documentclass[final,3p,times,twocolumn]{elsarticle}

\usepackage[numbers]{natbib}
\usepackage{graphicx}      
\usepackage[subfigure]{graphfig}
\usepackage{subfigure}
\usepackage{amsmath}
\usepackage{amssymb}
\usepackage{multirow}
\usepackage{float}
\usepackage{stfloats}
\usepackage{tabularx}
\usepackage{color}
\usepackage{algorithm}
\usepackage{algorithmic}
\usepackage{epstopdf}
\usepackage{makecell}
\usepackage{booktabs}

\begin{document}
	\renewcommand{\thefootnote}{\fnsymbol{footnote}}
	



\title{Monitoring multimode processes: a modified PCA algorithm with continual learning ability}

\author[label1]{Jingxin Zhang} 
\author[label2,label1]{Donghua Zhou \footnotemark[1]}  
\author[label1]{Maoyin Chen \footnotemark[1]}

\affiliation[label1]{organization={Depart of Automation, Tsinghua University},
	city={Beijing},
	postcode={100084},
	country={China}}
\affiliation[label2]{organization={College of Electrical Engineering and Automation, Shandong University of Science and Technology},
	city={Qindao},
	postcode={266000},
	country={China}}

%
%
%
%
%

\begin{abstract}
For multimode processes, one generally establishes local monitoring models corresponding to local modes. However, the significant features of previous modes may be catastrophically forgotten when a monitoring model for the current mode is built. It would result in an abrupt performance decrease.  It could be an effective manner to make local monitoring model remember the features of previous modes. Choosing the principal component analysis (PCA) as a basic monitoring model, we try to resolve this problem. A modified PCA algorithm is built with continual learning ability for monitoring multimode processes, which adopts elastic weight consolidation (EWC) to overcome catastrophic forgetting of PCA for successive modes. It is called PCA-EWC, where the significant features of previous modes are preserved when a PCA model is established for the current mode.  The optimal parameters are acquired by differences of convex functions. Moreover, the proposed PCA-EWC is extended to general multimode processes and the procedure is presented. The computational complexity and key parameters are discussed to further understand the relationship between PCA and the proposed algorithm. Potential limitations and relevant solutions are pointed to understand the algorithm further. Numerical case study and a practical industrial system in China are employed to illustrate the effectiveness of the proposed algorithm.
\end{abstract}



\maketitle

\section{Introduction}
Process monitoring is increasingly important due to the strict requirements for  safety and has received remarkable success  \cite{destro2020333A,ZHENG202010Enhanced,shang2018112Isolating,Shen2012A,zhang2019an,shi2020distributed}. 
In most industrial systems, operating condition often changes with raw materials, product specifications, maintenance, etc \cite{Huang2020structure,Zurita2018Multimodal,zhao2010statistical}.
Data from different modes have different characteristics, such as mean, covariance and distribution \cite{tan2020119An,song2016136key}. Thus, building an effective monitoring model for multimode processes is a challenging problem.  

Current research status for monitoring multimode processes is reviewed briefly.  Marcos \emph{et al}.  \cite{quiones-grueiro2019data} summarized researches for multimode processes and 
 divided the methods into two categories: 1) single models appropriate for every mode, where the multimodality features are removed by a transformation function and the monitoring model is built by the decision function in most cases \cite{ma2012a,deng2017nonlinear}; 2) multiple models where the mode is identified first and then local models are established in each mode \cite{wang2020data,xu2014multimode,peng2017multimode}.
 Ma \emph{el al}. \cite{ma2012a} utilized local neighborhood standardization strategy to transfer the multimode data into unimodal distribution approximately and only one model was constructed for process monitoring. The multimode data were transformed to probability density by local probability density estimation that obeyed unimodal distribution 
\cite{deng2017nonlinear}. However, the transformation function is difficult to be determined accurately for complex systems. Recursive methods are regarded as a single-model based method and can update the parameters when the modes change  \cite{Li2000Recursive}, which are appropriate for slow switching of working conditions.
Wang \emph{et al}. \cite{wang2020data} utilized Dirichlet process Gaussian mixed model to identify the mode automatically and then support vector data description was designed for each mode.
In \cite{xu2014multimode}, PCA mixture models was proposed to reduce the dimension and the number of mixture components was optimized automatically by Bayesian Ying-Yang algorithm. The performance of mixture models is greatly influenced by the accuracy of mode identification and data from all potential modes are required before training.

For current multimode monitoring methods, they generally require that training data must include all the modes. When two or more modes come sequentially, single model methods would train the model from scratch to acquire the transformation function based on data from all previous modes. However, it is intractable to obtain the appropriate transformation function.
For multiple models, local monitoring models are established corresponding to local modes. The important features of previous modes may be overwritten by new data when a single monitoring model is established for the current mode, thus delivering an abrupt decrease of performance. 
Consequently, we need to retrain the monitoring models using all the data  from scratch, 
which consumes huge storage space and computing resources. State-of-the-art methods generally extract critical features from collected data, which are
sufficient to reflect the data characteristics. 
Therefore, instead of storing original sensing data, is it possible to make local monitoring model remember the features of previous modes? Thus, a single model can show excellent performance for simultaneously monitoring the current mode and the future similar modes. For example, two  successive modes ${\mathcal{M}_1}$ and ${\mathcal{M}_2}$ need to be learned sequentially. If we build one model for the mode ${\mathcal{M}_2}$ based on the learned knowledge from the mode ${\mathcal{M}_1}$ and data from mode ${\mathcal{M}_2}$, the single model would preserve the features for both modes with a small loss and provide outstanding performance for multimode processes \cite{KirkpatrickOvercoming,parisi2019continual}. Once the mode ${\mathcal{M}_1}$ revisits, the single model can achieve acceptable monitoring results. 

In the field of artificial intelligence, continual learning is exactly the technique that trains the new model based on the new mode data and the partial information about previous modes \cite{parisi2019continual}.
However, continual learning remains a long-standing challenge owing to ``catastrophic forgetting'' issue, namely,  training a new model with new data would severely influence the performance of previous modes \cite{dachapally2018catastrophic,Aljundi2019Continual}.
Motivated by synaptic consolidation, Kirkpatrick \emph{et al}. proposed elastic weight consolidation (EWC) to overcome catastrophic forgetting issue \cite{KirkpatrickOvercoming}, which can learn consecutive tasks without forgetting previous tasks disastrously.

In this paper, we adopt the technique of EWC to overcome the catastrophic forgetting issue of PCA for multimode processes, referred to as PCA-EWC, where the significant features of previous modes are preserved if a PCA model is built for the current mode. An appropriate objective function is chosen to achieve continual learning, and the global optimal solution can be calculated by difference of convex functions (DC) \cite{zheng2009DC}.
The proposed PCA-EWC algorithm requires that similarity exists among different modes and it is easy to satisfy in practical applications because data in physical and chemical processes are generally governed by specific laws.
Besides, the proposed PCA-EWC algorithm is applied to multimode processes, if data are stationary for each mode and the operating condition can transform from stationary state to another stationary one quickly. Thus, the mode is identified by means as well as standard variances.

It is observed that the proposed PCA-EWC algorithm extracts relatively common significant information through updating the monitoring model continually when new modes arrive. Similarly, Zhao \emph{et al}. proposed statistical  analysis for multimode processes with between-mode transitions and extracted the common basis vectors shared by all modes, which spanned the mode-common subspace \cite{zhao2010statistical}.  It shares some common points with PCA-EWC: i) there exists a certain degree of similarity among modes; ii) the global basis vectors should be approximated to the optimal parameters for each mode. However, algorithm in \cite{zhao2010statistical} needs as complete data as possible and the transition pattern should be considered.  Literature \cite{zhao2015comprehensive} mainly focused on between-mode transition and decomposed the subspace based on the relative changes from the reference mode and effect on monitoring.  Different from aforementioned two methods, PCA-EWC is not required that data from multiple modes are available when training the  monitoring model. When a new mode arrives, the model is updated based on the previous model and the current data, thus delivering prominent performance for successive modes. PCA-EWC contains common features from the existing modes, which makes it possible to monitor the between-mode transition process. In this paper, the between-mode transition process is not considered  and will be investigated in future work.

The contributions of this paper are summarized as follows:

\begin{enumerate}[(a)]

\item It provides a novel framework PCA-EWC for monitoring multimode processes. The major merit is that significant features of previous modes are preserved when a PCA model is designed for the current mode;

\item Different from single model based methods, PCA-EWC updates the model using the learned knowledge without learning transformation function, which is appropriate for complex industrial systems;

\item Compared with multiple models, PCA-EWC preserves significant features from previous modes, and establishes a single model with continual learning ability that is effective for successive modes simultaneously.

\end{enumerate}

The remaining parts are organized below. Section \ref{EWC_THEORY} reviews the EWC algorithm and reformulates PCA from the probabilistic perspective briefly.
Section \ref{Sec:3} introduces PCA-EWC, the global optimal solution by DC programming, and the procedure for  monitoring multimode processes.
PCA-EWC is extended to more general multimode process monitoring and the specific  procedure is presented in Section IV.
Moreover, the relationship between PCA-EWC and PCA is discussed and the computational complexity is analyzed. A numerical case study and a practical plant subsystem are adopted to illustrate the effectiveness of the proposed algorithm in Section \ref{Sec:4}. The conclusion is given in Section \ref{Sec:5}.

\section{Preliminary}\label{EWC_THEORY}

In this section, we briefly introduce the core of EWC and PCA algorithms from the probabilistic perspective, and establish the foundation for the proposed PCA-EWC.
For convenience, we consider the successive monitoring tasks in successive modes  ${\mathcal M}_1$ and ${\mathcal M}_2$.

\subsection{The revisit of EWC}\label{theory_EWC}
EWC algorithm is an efficient method to overcome catastrophic forgetting issue \cite{KirkpatrickOvercoming}. It slows down the change on certain parameters based on the importance of previous tasks, in order to avoid dramatic performance degradation for previous similar modes.
For a learning method, learning a monitoring task principally adjusts the parameter $\theta$ by optimizing performance. Different configurations of $\theta$ may lead to the same result \cite{Sussmann1992Uniqueness}. This makes it possible that parameter of the latter mode ${\mathcal{M}_2}$, $\theta _{\mathcal{M}_2}^*$, is close to the parameter of previous mode $\mathcal{M}_1$, $\theta _{\mathcal{M}_1}^*$. Thus, when building the monitoring model for mode ${\mathcal{M}_2}$, partial information of mode ${\mathcal{M}_1}$ should be preserved,  and $\theta _{\mathcal{M}_1}^*$ and $\theta _{\mathcal{M}_2}^*$ have a certain degree of similarity.  

It is universally known that the learning processes are reformulated from the probabilistic perspective as follows. It is transformed into finding the most probable parameter given data set $\boldsymbol X$. According to Bayesian rule, the conditional probability is calculated by prior probability $p(\theta)$ and data probability $p(\boldsymbol X|\theta)$:
\begin{equation}\label{bayesian1}
	\log \,p\left( {\theta |\boldsymbol X} \right) = \log \,p\left( {\boldsymbol X|\theta } \right) + \log \,p\left( \theta  \right) - \log \,p\left( \boldsymbol X \right)
\end{equation}
Suppose that data $\boldsymbol X$ come from  two independent modes, namely, mode ${\mathcal{M}_1}$ ($\boldsymbol X_1$) and  mode ${\mathcal{M}_2}$ ($\boldsymbol X_2$).   According to Bayesian theory, (\ref{bayesian1}) can be reformulated as:
\begin{equation}\label{poster2}
	\log \,p\left( {\theta |\boldsymbol X} \right) = \log \,p\left( {\boldsymbol X_2|\theta } \right) + \log \,p\left( \theta |\boldsymbol X_1 \right) - \log \,p\left( \boldsymbol X_2 \right)
\end{equation}
Note that $  \log \,p\left( {\theta |\boldsymbol X} \right)  $ denotes the posterior probability of the parameters given the entire dataset. $-\log \,p\left( {\boldsymbol X_2|\theta } \right)$ represents the loss function for mode $ {\mathcal{M}_2} $. Posterior distribution $ \log \,p\left( \theta |\boldsymbol X_1 \right) $  can reflect all information of mode $ {\mathcal{M}_1} $ and significant information of mode $ {\mathcal{M}_1} $ is contained in the posterior probability.

The true posterior probability $p\left( \theta |\boldsymbol X_1 \right)$ is generally intractable to compute and  approximated by Laplace approximation in this paper \cite{MacKay1992A,huszar2017on}.
Detailed procedure has been presented in Appendix \ref{sec:2.2}.
Thus, the problem (\ref{poster2}) is transformed to (\ref{laplaceapproximation1}). The hyper-parameter $\lambda_{\mathcal{M}_1}$ also represents the importance of the previous mode.  Then, the purpose of EWC is to minimize
\begin{equation}\label{ewc_final}
	\begin{aligned}
		- \log \,p\left( {\theta |\boldsymbol X} \right) \approx & -\log \,p\left( {\boldsymbol X_2|\theta } \right) +\frac{1}{2}(\theta-{\theta _{{\mathcal{M}_1}}^*})^T \\
		& (\lambda_{\mathcal{M}_1} \boldsymbol F_{{\mathcal{M}_1}}+\lambda_{prior} \boldsymbol I)(\theta-{\theta _{{\mathcal{M}_1}}^*})
	\end{aligned}
\end{equation}
Let
\begin{equation}\label{semidefinitematrix}
	\boldsymbol \Omega_{{\mathcal{M}_1}}= \frac{1}{2}(\lambda_{\mathcal{M}_1} \boldsymbol F_{{\mathcal{M}_1}}+\lambda_{prior} \boldsymbol I)
\end{equation}
The objective (\ref{ewc_final}) is simplified by
\begin{equation}\label{objective_ewc}
	\begin{aligned}
		\mathcal{J}(\theta) =
		\mathcal{J}_2(\theta,\boldsymbol X_2)+   \mathcal{J}_{loss}(\theta, {\theta _{{\mathcal{M}_1}}^*},\boldsymbol \Omega_{{\mathcal{M}_1}})
	\end{aligned}
\end{equation}
where $\mathcal{J}_2(\theta,\boldsymbol X_2) = -\log \,p\left( {\boldsymbol X_2|\theta } \right)$  represents the loss function of $\boldsymbol X_2$.
$\mathcal{J}_{loss} = (\theta-{\theta _{{\mathcal{M}_1}}^*})^T  \boldsymbol \Omega_{{\mathcal{M}_1}}(\theta-{\theta _{{\mathcal{M}_1}}^*})$ reflects the total loss of previous modes and measures the disparity between the last mode and the current mode. 
 Note that we discard $\boldsymbol X_1$ after learning the model for mode ${\mathcal{M}_1}$. The recent model (\ref{objective_ewc}) is built based on   $\boldsymbol X_2$ and the parameters from   mode $ \mathcal{M}_1 $, without the requirement of $\boldsymbol X_1$.

\subsection{Probabilistic perspective of PCA}\label{probabilisticPCA}
The observation data $\boldsymbol x \in R^{m}$ are generated from latent variables $\boldsymbol y \in R^{l}$ with $\boldsymbol y \sim N(\boldsymbol 0,\boldsymbol I)$, then
\begin{equation}\label{PCA1}
	\boldsymbol {x} = {\boldsymbol P} {\boldsymbol y}  + \boldsymbol \mu+ {\boldsymbol \xi}
\end{equation}
where $\boldsymbol \mu \in R^{m}$ is the mean value, noise $\boldsymbol \xi_i \sim N(0,\sigma^2)$ , $i=1,\cdots,m$, ${\sigma ^2}$ is constant but unknown,   $\boldsymbol P \in R^{m \times l}$ is the loading matrix. Thus, the conditional probability is \cite{zhang2019an,tipping1999mixtures}
\begin{equation}\label{eqxy}
	p({\boldsymbol x}|{\boldsymbol y}) = {(2\pi {\sigma ^2})^{ -\frac{m}{2}}}\exp \left\{ { - \frac{1}{{2{\sigma ^2}}}{{\left\| {\boldsymbol x - {\boldsymbol P} {\boldsymbol y} - \boldsymbol \mu } \right\|}_2^2}} \right\}
\end{equation}
The Gaussian prior of latent variables is
\begin{equation}\label{eqy}
	p({\boldsymbol y}) = {(2\pi)^{ -\frac{l}{2}}}\exp \left\{ { - \frac{1}{{2}}{{\left\| {\boldsymbol y } \right\|}_2^2}} \right\}
\end{equation}
Based on Beysian theory, the marginal probability is
\begin{equation}
	\begin{aligned}
		&	p\left( {{\boldsymbol x_i},{\boldsymbol y_i}} \right) 
		= {(2\pi)^{ -\frac{m+l}{2}}} {\sigma}^{ -{m}} \exp \left\{ { - \frac{1}{{2{\sigma ^2}}}{{\left\| {\boldsymbol x_i - {\boldsymbol P} {\boldsymbol y_i} - \boldsymbol \mu } \right\|}_2^2}- \frac{1}{{2}}{{\left\| {\boldsymbol y_i } \right\|}_2^2}} \right\} 
	\end{aligned}
\end{equation}
The log-likelihood of observed data is \cite{zhang2019an,tipping1999mixtures}
\begin{equation}\label{likelihood1}
	L = \sum\nolimits_i {\ln \left\{ {p\left( {{\boldsymbol x_i},{\boldsymbol y_i}} \right)} \right\}}
\end{equation}
Thus, the expectation of $L$ is
\begin{equation} \label{likelihood2}
	\begin{aligned}
		\left\langle L \right\rangle
		= & - \frac{{m + l}}{2}\ln \;2\pi  - m\ln \,\sigma  - \frac{1}{{2{\sigma ^2}}} \sum\nolimits_i {\left\| {\boldsymbol x_i - {\boldsymbol P} {\boldsymbol y_i} - \boldsymbol \mu} \right\|}_2^2\\
		&	- \frac{1}{2} \sum\nolimits_i {\left\| {\boldsymbol y_i } \right\|}_2^2
	\end{aligned}
\end{equation}
When $\sigma  \to 0$, probabilistic PCA is transformed to PCA, and maximization of (\ref{likelihood2}) is equivalent to minimize
\begin{equation}\label{posterior2}
	\begin{aligned}
	\sum\limits_{i = 1}^N {\left\| {\boldsymbol x_i - {\boldsymbol P} {\boldsymbol y_i} - \boldsymbol \mu} \right\|}_2^2
	= tr(\bar{\boldsymbol X}^T \bar{\boldsymbol X})
\end{aligned}
\end{equation}
where $\bar{\boldsymbol X} =\boldsymbol X- \boldsymbol 1 \boldsymbol \mu -\boldsymbol Y \boldsymbol P^T$, $\boldsymbol Y = [\boldsymbol y_1, \cdots, \boldsymbol y_N]$,  $\boldsymbol P^T \boldsymbol P = \boldsymbol I$, $\boldsymbol 1$ is the vector of all ones with appropriate dimension, $N$ is the number of samples.
According to (\ref{PCA1}),  $\boldsymbol Y = \boldsymbol X \boldsymbol P -\boldsymbol 1 \boldsymbol \mu \boldsymbol P-\boldsymbol \Xi$, $\boldsymbol \Xi$ is the noise. Thus, $\bar{\boldsymbol X}= (\boldsymbol X- \boldsymbol 1 \boldsymbol \mu)(\boldsymbol I -\boldsymbol P \boldsymbol P^T) - \boldsymbol \Xi  \boldsymbol P $.

\subsection{The foundation of PCA-EWC}\label{foundationPCAEWC}
This section introduces the association between PCA and EWC, which lays the solid foundation of the proposed PCA-EWC.
The major problem is to acquire the specific reformulation of (\ref{objective_ewc}) based on PCA.
Thus, we need to calculate each term of right-hand side for (\ref{objective_ewc}).

Based on (\ref{likelihood1},\ref{posterior2}), the first term of (\ref{objective_ewc}) is designed as
\begin{equation}\label{pca_posterior}
	\mathcal{J}_2(\boldsymbol P,\boldsymbol X)
	= tr((\boldsymbol I -\boldsymbol P \boldsymbol P^T)(\boldsymbol X- \boldsymbol 1 \boldsymbol \mu)^T(\boldsymbol X- \boldsymbol 1 \boldsymbol \mu))\\
\end{equation}

For the second term $\mathcal{J}_{loss}$, 
the key is to determine $\boldsymbol \Omega_{\mathcal{M}_1}$ by  (\ref{semidefinitematrix}), which is actually the second deviation of (\ref{likelihood1}) in Appendix \ref{sec:2.2}. 
Therefore,  the detailed form of (\ref{objective_ewc}) is acquired based on PCA.

\section{PCA-EWC for monitoring  multimode processes}\label{Sec:3}
In this section, we present the procedure of PCA-EWC with continual learning ability and use DC programming to obtain the optimal solution, which is then applied to monitor multimode processes. Here, we consider monitoring tasks for two successive modes.%

\subsection{Problem formulation}\label{Sec:3.0}
Assume that there exist two sequential monitoring tasks corresponding to successive modes ${\mathcal{M}_1}$ and ${\mathcal{M}_2}$. The basic PCA is used to build a monitoring model.
Given normal data  ${{\boldsymbol {X}}_1 \in R^{N_1\times m}}$, ${\boldsymbol X}_2\in R^{N_2 \times m}$, ${N_i}\; (i=1,2)$ and $m$ are the number of samples and variables, respectively. For brevity, data are already scaled to zero mean and unit variance. 

With regard to the monitoring task for the mode ${\mathcal M}_1$, the projection matrix is ${\boldsymbol  P_{{\mathcal M}_1} \in R^{m \times l}}$ through PCA. $l$ is the number of principal components determined by cumulative percent variance approach and the threshold is set to be $0.95$ in this paper. Thus, the purpose is to find a proper projection matrix $\boldsymbol {P}$, which is effective to  monitor modes ${\mathcal M}_1$ and ${\mathcal M}_2$ simultaneously.

\begin{figure}[!tbp]
	\centering
	\includegraphics[width=0.45\textwidth]{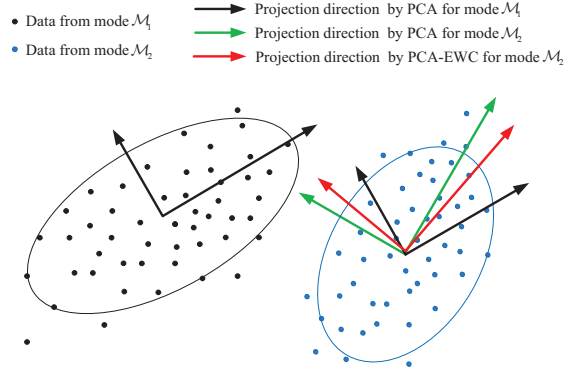}
	\caption{Geometric illustration of EWC-PCA}
	\label{fig_theory}
\end{figure}

As illustrated in Figure \ref{fig_theory}, after monitoring task in the mode ${\mathcal M}_1$ is learned by PCA, the parameter is at $\boldsymbol  P_{{\mathcal M}_1}$ (the black arrow). If we train the monitoring task for the mode ${\mathcal M}_2$ alone (the green arrow), the learned knowledge from the mode ${\mathcal M}_1$ would be destroyed or even completely overwritten by new data in the worst case. EWC enables us to monitor the mode ${\mathcal M}_2$ without suffering important loss on the mode ${\mathcal M}_1$ (the red arrow), by preserving the partial information for the mode ${\mathcal M}_1$.

\subsection{PCA-EWC algorithm}\label{Sec:3.1}

For simplicity, we assume that $l$ remains the same for two successive modes.
According to (\ref{objective_ewc}), the objective function is described as:
\begin{equation}\label{total_loss}
	\mathcal{J}(\boldsymbol {P}) = \mathcal{J}_2(\boldsymbol {P})+ \mathcal{J}_{loss}(\boldsymbol {P},\boldsymbol  P_{{\mathcal M}_1},{\boldsymbol \Omega}_{{\mathcal M}_1})
\end{equation}
where $\mathcal{J}_2(\boldsymbol {P})$ is the loss function for monitoring task in the mode ${\mathcal M}_2$ only, 
$\mathcal{J}_{loss}(\boldsymbol P,\boldsymbol  P_{{\mathcal M}_1},{\boldsymbol \Omega}_{{\mathcal M}_1}) $ represents the loss function of previous mode $ {{\mathcal M}_1} $, and $ {\boldsymbol \Omega}_{{\mathcal M}_1}  $ is positive definite and calculated by (\ref{semidefinitematrix}).

For PCA, 
according to (\ref{pca_posterior}),  $\mathcal{J}_2(\boldsymbol {P})$  is reformulated as
\begin{equation}\label{loss1}
	\mathcal{J}_2(\boldsymbol {P})
	= tr(\boldsymbol X_2^T \boldsymbol X_2)-tr(  \boldsymbol  P^T  \boldsymbol X_2^T \boldsymbol X_2 \boldsymbol  P)
\end{equation}
where the constraint $\boldsymbol  P^T \boldsymbol  P = \boldsymbol I$ with ${\boldsymbol {P} \in R^{m \times l}}$, $ \boldsymbol \mu = \boldsymbol 0$.

Based on (\ref{ewc_final}-\ref{objective_ewc}),
$\mathcal{J}_{loss}(\boldsymbol {P},\boldsymbol  P_{{\mathcal M}_1},{\boldsymbol \Omega}_{{\mathcal M}_1}) $ is expressed by:
\begin{equation}\label{loss2}
	\begin{aligned}
		&\mathcal{J}_{loss}(\boldsymbol {P},\boldsymbol  P_{{\mathcal M}_1},{\boldsymbol \Omega}_{{\mathcal M}_1}) \\
		=& tr\{(\boldsymbol  P - \boldsymbol  P_{{\mathcal M}_1})^T {\boldsymbol \Omega}_{{\mathcal M}_1} (\boldsymbol  P - \boldsymbol  P_{{\mathcal M}_1} )\} \\
		=&\|\boldsymbol {P}-\boldsymbol {P}_{\mathcal{M}_1}\|_{{\boldsymbol \Omega}_{\mathcal{M}_1}}^2
	\end{aligned}
\end{equation}
Obviously, $ \mathcal{J}_{loss}(\boldsymbol {P},\boldsymbol  P_{{\mathcal M}_1},{\boldsymbol \Omega}_{{\mathcal M}_1}) $ 
measures the difference of current parameter $ \boldsymbol {P}_{\mathcal{M}_1} $ and the optimal parameter $ \boldsymbol {P} $.

Substituting (\ref{loss1}-\ref{loss2}) into (\ref{total_loss}), we can get
\begin{equation}\label{objective_original}
	\begin{aligned}
		\mathcal{J}(\boldsymbol {P}) 
		=& tr(\boldsymbol  P^T {\boldsymbol \Omega}_{{\mathcal M}_1} \boldsymbol P)-tr(\boldsymbol  P^T  \boldsymbol X_2^T \boldsymbol X_2 \boldsymbol  P) -2  tr(\boldsymbol  P^T {\boldsymbol \Omega}_{{\mathcal M}_1} \boldsymbol  P_{{\mathcal M}_1}) \\
		& + \underbrace{\{tr(\boldsymbol X_2^T \boldsymbol X_2)+ tr(\boldsymbol  P_{{\mathcal M}_1}^T {\boldsymbol \Omega}_{{\mathcal M}_1} \boldsymbol  P_{{\mathcal M}_1})\}}_{constant}
	\end{aligned}
\end{equation}

Problem (\ref{objective_original}) is nonconvex and  intractable to acquire the global optimal solution by stochastic gradient descent method.
Let $G(\boldsymbol P) =  tr(\boldsymbol  P^T {\boldsymbol \Omega}_{{\mathcal M}_1} \boldsymbol P)-2  tr(\boldsymbol  P^T {\boldsymbol \Omega}_{{\mathcal M}_1} \boldsymbol  P_{{\mathcal M}_1})$, $H(\boldsymbol P) = tr(\boldsymbol  P^T  \boldsymbol X_2^T \boldsymbol X_2 \boldsymbol  P)$. Thus, $\mathcal{J}(\boldsymbol {P}) = G(\boldsymbol P)-H(\boldsymbol P)+constant $.
The original problem (\ref{total_loss}) can be transformed into  
\begin{equation}\label{objective_function}
	\begin{aligned}
		& \mathop{min}\limits_{\boldsymbol P} \quad \mathcal{J}(\boldsymbol {P}) \Longleftrightarrow \mathop{min}\limits_{\boldsymbol P} \quad G(\boldsymbol P)-H(\boldsymbol P) \\
		& s.t. \qquad \boldsymbol  P^T \boldsymbol  P = \boldsymbol I \in R^{l \times l}
	\end{aligned}
\end{equation}

As $G(\boldsymbol P)$ and $H(\boldsymbol P)$ are convex,  the objective function (\ref{objective_function}) is formulated as DC programming problem \cite{Souza2015Global}.

\renewcommand{\algorithmicrequire}{ \textbf{Input:}} 
\renewcommand{\algorithmicensure}{ \textbf{Output:}} 

\begin{algorithm}[!bp]
	\caption{PCA-EWC using DC programming.}\label{ALG1}
	\begin{algorithmic}[1]
		\REQUIRE  parameter $\epsilon$ \\ 
		\ENSURE  $\boldsymbol P_{k+1}$, that yields the minimum of $\mathcal{J}(\boldsymbol {P})= G(\boldsymbol P)-H(\boldsymbol P)$ with constraint $\boldsymbol P^T \boldsymbol P = \boldsymbol I $ \\
		\STATE  Let $\boldsymbol P_0 = \boldsymbol  P_{{\mathcal M}_1}$ be an initial solution\\
		\STATE  Set the initial counter $k=0$
		\STATE Compute $\boldsymbol P_{k+1}$ by solving the problem (\ref{final_objective2})
		\STATE Calculate $ \boldsymbol P_{k+1} = \boldsymbol W_k \boldsymbol I_{m,l} \boldsymbol V_k^T $
		\STATE Let $k=k+1$, go to step 3 until  $ \| \boldsymbol P_{k+1} - \boldsymbol P_k\|_F^2 < \epsilon$  
	\end{algorithmic}
\end{algorithm}

\begin{table*}[htbp]
	\begin{center}
		\caption{Simulation scheme of PCA-EWC}\label{Table-simulation}
		\footnotesize
		\begin{tabular}{c c c c c}
			\hline
			&    \makecell{Training resources}    & \makecell{Training model label}   & \makecell{Algorithm} & \makecell{Testing data} \\
			\hline
			Situation 1 & Training data 1            & Model A    & PCA      & Testing data 1 \\
			Situation 2 & Training data 2 + Model A  & Model B    & PCA-EWC  & Testing data 2\\
			Situation 3 &  -                         & Model B    & -        & Testing data 3 \\
			Situation 4 & Training data 2            & Model C    & PCA      & Testing data 3\\
			\hline
		\end{tabular}
	\end{center}
\end{table*}

\subsection{Optimal solution based on DC programming}\label{Sec:3.2}
Inspired by solution in \cite{nguyen2018approach}, we adopt DC programming to  optimize (\ref{objective_function}), as summarized in  Algorithm \ref{ALG1}. The procedure includes linearizing the concave part and solving the convex subproblem part.

\subsubsection{Linearizing the concave part} 
Assume that $\boldsymbol P_k$ is the solution at $k$th iteration in Algorithm \ref{ALG1}.
The linearization of  $H(\boldsymbol P)$  is given by
\begin{center}
	$H_l(\boldsymbol P) = H(\boldsymbol P_k)+\langle \boldsymbol P-\boldsymbol P_k, \boldsymbol U_k\rangle, \boldsymbol U_k \in \partial H(\boldsymbol P_k)$
\end{center}
Then, (\ref{objective_function}) can be approximated by solving a convex program since $H_l(\boldsymbol P)$ is a linear function of $ \boldsymbol P$.  In order to approximate the concave part, we need to compute the subgradient $ \boldsymbol U_k$. 
Since $\boldsymbol U \in \dfrac{\partial H(\boldsymbol P)}{ \partial \boldsymbol P} = 2 \boldsymbol X_2^T \boldsymbol X_2 \boldsymbol P$, let $\boldsymbol U_k = 2\boldsymbol X_2^T \boldsymbol X_2 \boldsymbol P_k$.

\subsubsection{Solving the convex subproblem}

After obtaining a subgradient $\boldsymbol U_k$ of $H(\boldsymbol P)$ at $\boldsymbol P_k$,  $H(\boldsymbol P)$ is replaced by its linearization. Therefore, (\ref{objective_function}) is approximated by  
\begin{equation}\label{objective_subproblem}
	\boldsymbol P_{k+1} \doteq \underset{\boldsymbol P^T \boldsymbol P = \boldsymbol I}{\arg \min} \quad G(\boldsymbol P)-<\boldsymbol P,\boldsymbol U_k>
\end{equation}
Since ${\boldsymbol \Omega}_{{\mathcal M}_1}$ is positive definite, let ${\boldsymbol \Omega}_{{\mathcal M}_1} = \boldsymbol L^T \boldsymbol L$ and $\boldsymbol L$ is the triangle matrix. Thus, we can get
\begin{equation}\label{formulated_solution}
	\begin{aligned}
		& G(\boldsymbol P)-<\boldsymbol P,\boldsymbol U_k> \\
		=& tr(\boldsymbol P^T{\boldsymbol \Omega}_{{\mathcal M}_1}\boldsymbol P)-2tr(\boldsymbol P^T {\boldsymbol \Omega}_{{\mathcal M}_1} \boldsymbol  P_{{\mathcal M}_1})-2tr(\boldsymbol P^T \boldsymbol X_2^T \boldsymbol X_2 \boldsymbol P_k)\\
		= & {\langle \boldsymbol L\boldsymbol P, \boldsymbol L\boldsymbol P\rangle}-2 \langle \boldsymbol L\boldsymbol P, \boldsymbol L\boldsymbol  P_{{\mathcal M}_1}+(\boldsymbol L^T)^{-1}\boldsymbol X_2^T \boldsymbol X_2 \boldsymbol P_k  \rangle \\
		=& \lVert  \boldsymbol Z_k- \boldsymbol L\boldsymbol P\rVert _F^2- \lVert  \boldsymbol Z_k\rVert _F^2
	\end{aligned}
\end{equation}
where $ \boldsymbol Z_k = \boldsymbol L\boldsymbol  P_{{\mathcal M}_1}+(\boldsymbol L^T)^{-1}\boldsymbol X_2^T \boldsymbol X_2 \boldsymbol P_k $, and it is constant at $k+1$th iteration.

Then, the optimization problem (\ref{objective_subproblem}) is equivalent to
\begin{equation}\label{final_objective1}
	\boldsymbol P_{k+1} = \underset{\boldsymbol P^T \boldsymbol P = \boldsymbol I}{\arg \min} \quad \lVert  \boldsymbol Z_k- \boldsymbol L\boldsymbol P\rVert _F^2
\end{equation}
Motivated by section 3.5 in \cite{huang2014robust}, (\ref{final_objective1}) is reformulated as
\begin{equation}\label{final_objective2}
	\boldsymbol P_{k+1} = \underset{\boldsymbol P^T \boldsymbol P = \boldsymbol I}{\arg \min} \quad \lVert  \boldsymbol P-\boldsymbol L^{T} \boldsymbol Z_k\rVert _F^2
\end{equation}
Let $\boldsymbol R_k=\boldsymbol L^{T} \boldsymbol Z_k= {\boldsymbol \Omega}_{{\mathcal M}_1} \boldsymbol  P_{{\mathcal M}_1}+\boldsymbol X_2^T \boldsymbol X_2 \boldsymbol P_k$. According to the lemma in \cite{huang2014robust},  $ \boldsymbol P_{k+1} = \boldsymbol W_k \boldsymbol I_{m,l} \boldsymbol V_k^T $, where $\boldsymbol W_k \in R^{m \times m}$ and $\boldsymbol V_k \in R^{l \times l}$ are  left and right singular vectors of the singular vector decomposition (SVD) of $\boldsymbol R_k$. Detailed  derivation process can be found in \cite{huang2014robust}.
Notice that the optimal projection matrix for the mode ${\mathcal M}_2$ is denoted as $\boldsymbol P_{{\mathcal M}_2}$ for convenience.

\begin{algorithm}[!bp]
	\caption{Off-line training phase}\label{off-line}
	\begin{algorithmic}[1]
		\STATE  For Situation 1, calculate the mean $\boldsymbol  \mu_1$ and variance $\boldsymbol  \Sigma_1$ of training data 1  and normalize data;
		
		\STATE    Train PCA model using training data 1 and calculate $\boldsymbol  P_{{\mathcal M}_1}$. The  model is denoted as Model A;
		
		\STATE Calculate mean $\boldsymbol  \mu_2$ and variance  $\boldsymbol  \Sigma_2$ of training data 2 and normalize data;
		
		\STATE   Train the new mode ${\mathcal M}_2$ by PCA-EWC, and calculate the optimal $\boldsymbol P_{{\mathcal M}_2}$ by Algorithm \ref{ALG1}. This training model is recorded as Model B;
		
		\STATE   For Situation 4, train the monitoring model for the mode ${\mathcal M}_2$ by PCA, labeled by Model C;
		
		\STATE For Situations 1, 2 and 4, calculate test statistics by (\ref{calculateT2}-\ref{calculatespe1}), and compute thresholds by KDE;
		
		
	\end{algorithmic}
\end{algorithm}

\subsection{Summary}\label{Sec:3.3}

Similar to many machine learning methods, two test statistics are designed for process monitoring. Hotelling's $T^2$ is calculated to monitor the principal component subspace and squared prediction error (SPE) is calculated for the residual component subspace.
\begin{equation}\label{calculateT2}
	{T^2} = \boldsymbol x \boldsymbol P{\left( {\frac{{{\boldsymbol P^T}{\boldsymbol X^T}\boldsymbol X \boldsymbol P}}{{N - 1}}} \right)^{ - 1}}{\boldsymbol P^T}{\boldsymbol x^T}
\end{equation}
\begin{equation}\label{calculatespe1}
	{\rm SPE} = \boldsymbol x (\boldsymbol I-\boldsymbol P \boldsymbol P^T)\boldsymbol x^T
\end{equation}
Note that $\boldsymbol P=\boldsymbol  P_{{\mathcal M}_i}$, $\boldsymbol X = \boldsymbol X_i$ and $N=N_i$ for the mode ${\mathcal M}_i$.

Given a confidence limit $\alpha$, the thresholds of  two statistics are calculated by kernel density estimation (KDE) \cite{zhang2019an} and denoted as $J_{th,T^2}$ and $J_{th,\rm SPE}$.  Note $\alpha = 0.99$ in this paper. Thus, the  detection logic satisfies\\
${T^2} \le {J_{th,{T^2}}}$ and   $ SPE \le {J_{th, \rm SPE}}$  $\Rightarrow $ fault free, otherwise faulty.

We strive to highlight the continual learning ability of PCA-EWC for monitoring multimode processes, and illustrate the memory characteristic of current model for previous or similar modes. We set four combinations of training  and testing data to interpret the effectiveness of PCA-EWC, as depicted in Table \ref{Table-simulation}. Data 1 are from the mode ${\mathcal M}_1$  and Data 3  follow  basically the same or similar distribution. Data 2 are originated from the mode ${\mathcal M}_2$. Data 1, Data 2 and Data 3 are collected  successively. For each Data $i$, $i=1,2,3$, normal training data and the corresponding testing data are denoted as training data $i$  and testing data $i$. Notice that Situations 1 and 2 must be learned sequentially. For convenience,  we recur to the information in Table \ref{Table-simulation} to summarize the monitoring procedure. The off-line training and on-line monitoring phases are summarized in Algorithm \ref{off-line} and  Algorithm \ref{on-line}.

\begin{algorithm}[!tp]
	\caption{On-line monitoring phase}\label{on-line}
	\begin{algorithmic}[1]
		\STATE  For Situation 1, preprocess the testing data 1 by $\boldsymbol  \mu_1$ and $\boldsymbol  \Sigma_1$, and then utilize Model A to calculate two test statistics  in (\ref{calculateT2}-\ref{calculatespe1});
		
		\STATE    For Situation 2, preprocess the testing data 2 by $\boldsymbol  \mu_2$ and $\boldsymbol  \Sigma_2$, and employ Model B to calculate two test statistics;
		
		\STATE  For Situation 3, calculate mean $\boldsymbol  \mu_3$ and variance $\boldsymbol  \Sigma_3$ using normal data 3, preprocess test data 3 and adopt Model B  to calculate test statistics by (\ref{calculateT2}-\ref{calculatespe1});
		
		\STATE   For Situation 4, preprocess the testing data 3 by $ \boldsymbol  \mu_3$ and $\boldsymbol  \Sigma_3$, apply Model C  to calculate test statistics;
		
		\STATE   Defect faults according to the fault detection logic.
	\end{algorithmic}
\end{algorithm}

We assume that the process works under the normal condition at stable initial stage when  a new operating mode appears. Thus, mean and variance are calculated by a few normal data of the new mode and then utilized to preprocess the corresponding testing data.
Two indexes are adopted to evaluate the performance, namely, fault detection rate (FDR) and false alarm rate (FAR). The calculation method refers to \cite{zhang2019an}. Furthermore,  detection delay (DD) is valuable and the primary evaluation indicator for practical industrial systems. The detection delay refers to the number of samples that the fault is detected later than the practical abnormal time. 

\newdefinition{rmk}{Remark}
\begin{rmk}
  Model B is built by the significant features from the mode ${\mathcal M}_1$ and new data from the mode ${\mathcal M}_2$. If the performance of Situation 3 is excellent, the PCA-EWC model overcomes catastrophic forgetting issue and partial information of previous modes is sufficient to provide favorable capability. When the performance of Situations 2 and 3 is similarly excellent, it illustrates that PCA-EWC is effective to monitor the current mode and the future similar modes. Situation 4 is designed as a comparative study. If the monitoring effect of Situation 4 is poor, it illustrates that traditional PCA suffers from catastrophic forgetting issue and fails to detect novelty for multimode processes.
\end{rmk}

\section{Model extension and discussion}

\subsection{Model Extension}
We discuss three successive modes for process monitoring and  give the more general procedure briefly.

When data $ \boldsymbol X_3  $ from mode $\mathcal{M}_3$ are collected and the previous data are not accessible, the Bayesian posterior is decomposed below:
\begin{equation}\label{thirdmode}
	\log \,p\left( {\theta |\boldsymbol X} \right) = \log \,p\left( {\boldsymbol X_3|\theta } \right) + \log \,p\left( \theta |\boldsymbol X_1, \boldsymbol X_2 \right) + constant
\end{equation}
Here,  $ \boldsymbol X $ contain data from three modes. After learning the model from mode $\mathcal{M}_1$, data  $ \boldsymbol X_1 $ are discarded.
We adopt recursive Laplace approximation \cite{huszar2017on} to approximate (\ref{thirdmode}), as presented in  Appendix \ref{RecursiveLA}.

 Similar to (\ref{objective_ewc}), the objective function is designed as:
  \begin{equation}\label{objective_third}
  \mathcal{J}(\theta) = \mathcal{J}_3(\theta,\boldsymbol X_3)+   \mathcal{J}_{loss}(\theta, {\theta _{{\mathcal{M}_2}}^*},\boldsymbol \Omega_{{\mathcal{M}_2}})
  \end{equation}

Then, the above derivation is extended to more general cases. When a new mode $\mathcal{M}_n$ appears and needs to be learned, the data collected before are discarded. Let the data denote as $\boldsymbol X_n$, the objective is
\begin{equation}\label{moremode}
	\begin{aligned}
		\log \,p\left( {\theta |\boldsymbol X} \right) =& \log \,p\left( {\boldsymbol X_n|\theta } \right) + \log \,p\left( \theta |\boldsymbol X_1, \cdots, \boldsymbol X_{n-1} \right) \\
		 & + constant
	\end{aligned}
\end{equation}

Similarly,  (\ref{moremode}) is approximated by recursive Laplace approximation and reformulated as:
\begin{equation}\label{ewc_final_moremode}
	\begin{aligned}
		-\log \,p\left( {\theta |\boldsymbol X} \right) \approx& -\log \,p\left( {\boldsymbol X_n|\theta } \right) +(\theta-{\theta _{{\mathcal{M}_{n-1}}}^*})^T \\
		& {\boldsymbol \Omega}_{\mathcal{M}_{n-1}} (\theta-{\theta _{{\mathcal{M}_{n-1}}}^*})+constant
	\end{aligned}
\end{equation}
where
\begin{equation}
 {\boldsymbol \Omega}_{\mathcal{M}_{n-1}} = {\boldsymbol \Omega}_{\mathcal{M}_{n-2}}+ \frac{1}{2}\lambda_{\mathcal{M}_{n-1}} \boldsymbol F_{\mathcal{M}_{n-1}}, n \geq 3
\end{equation}
$\boldsymbol F_{\mathcal{M}_{n-1}}$ is the Fisher information matrix of mode $\mathcal{M}_{n-1}$,   $\lambda_{\mathcal{M}_{n-1}}$ is the hyper-parameter that measures the importance of the previous modes.
Then, the objective function is designed as
\begin{equation}\label{total_lossC}
	\mathcal{J}(\boldsymbol {P}) = \mathcal{J}_n(\boldsymbol {P})+ \mathcal{J}_{loss}(\boldsymbol {P},\boldsymbol {P}_{\mathcal{M}_{n-1}},{{\boldsymbol \Omega}_{\mathcal{M}_{n-1}}})
\end{equation}
where $\boldsymbol {P}_{\mathcal{M}_{n-1}}$ is the optimal projection matrix based on PCA-EWC for the previous mode $\mathcal{M}_{n-1}$.
$\mathcal{J}_n(\boldsymbol {P})$ is the loss function for the mode $\mathcal{M}_{n}$ by PCA. $	\mathcal{J}_{loss}(\boldsymbol {P},\boldsymbol {P}_{\mathcal{M}_{n-1}},{\boldsymbol \Omega_{n-1}})$ is a surrogate loss that approximates the total loss of previous $n-1$ modes. Similarly,
\begin{equation}\label{lossC}
	\mathcal{J}_n(\boldsymbol {P})
	= tr(\boldsymbol X_n^T \boldsymbol X_n)-tr(  \boldsymbol  P^T  \boldsymbol X_n^T \boldsymbol X_n \boldsymbol  P)
\end{equation}
\begin{equation}
	\mathcal{J}_{loss}(\boldsymbol {P},\boldsymbol {P}_{\mathcal{M}_{n-1}},{\boldsymbol \Omega_{n-1}}) = \|\boldsymbol {P}-\boldsymbol {P}_{\mathcal{M}_{n-1}}\|_{{\boldsymbol \Omega}_{\mathcal{M}_{n-1}}}^2
\end{equation}
Hence,
\begin{equation}\label{objective_originalC}
	\begin{aligned}
		\mathcal{J}(\boldsymbol {P}) =&  tr(\boldsymbol  P^T {{{\boldsymbol \Omega}_{\mathcal{M}_{n-1}}}} \boldsymbol P)-2  tr(\boldsymbol  P^T {\boldsymbol \Omega_{\mathcal{M}_{n-1}}} \boldsymbol  P_{{\mathcal M}_{n-1}}) \\
		&-tr(\boldsymbol  P^T  \boldsymbol X_n^T \boldsymbol X_n \boldsymbol  P)\\
		& + \underbrace{\{tr(\boldsymbol X_n^T \boldsymbol X_n)+ tr(\boldsymbol  P_{{\mathcal M}_{n-1}}^T {\boldsymbol \Omega_{\mathcal{M}_{n-1}}} \boldsymbol  P_{{\mathcal M}_{n-1}})\}}_{constant}
	\end{aligned}
\end{equation}

Minimization of (\ref{objective_originalC}) is settled by DC programming in  Section \ref{Sec:3.2}, denoted as $\boldsymbol {P}_{{\mathcal M}_n}$. The monitoring scheme can refer to Section \ref{Sec:3.3}.

\subsection{Discussion}

\subsubsection{The influence of parameter $\lambda$}

The matrix ${\boldsymbol \Omega}$ is influenced by $\lambda$. Large value of $\lambda$  indicates that the previous mode would play a significant role and  more information is expected to be retained. Generally, $\lambda$ is determined by prior knowledge.

Two extreme cases are described to strength understanding and we adopt Figure \ref{fig_theory} to illustrate specifically. When ${\lambda_{{\mathcal M}_1} = 0}$, the information of previous mode ${{\mathcal M}_1}$ is completely forgotten. PCA-EWC is equivalent to standard PCA and the training parameter of the mode ${\mathcal M}_2$ is shown by the green arrow in Figure \ref{fig_theory}.
When ${\lambda_{{\mathcal M}_1}  \to \infty} $, all information of previous mode is retained and the information from the current mode is nearly neglected. Thus, the training parameter of the mode ${\mathcal M}_2$ approximates to $\boldsymbol  P_{{\mathcal M}_1}$ (the dark arrow in Figure \ref{fig_theory}).

\subsubsection{Computational complexity analysis}

The computational complexity mainly contains training models of the mode ${\mathcal M}_1$ by PCA and the mode ${\mathcal M}_2$ by PCA-EWC. In Algorithm \ref{ALG1}, the computation focuses on the SVD of $\boldsymbol R_k \in R^{m \times l}$.
The term \emph{flam} is utilized to measure the operation counts.
The SVD of $\boldsymbol X_1$ needs $\dfrac{3}{2}m^2N_1+\dfrac{9}{2}m^3$ flam. For the  monitoring task from the mode ${\mathcal M}_2$,
the matrices ${\boldsymbol \Omega}_{{\mathcal M}_1} \boldsymbol  P_{{\mathcal M}_1} \in R^{m \times l}$ and  $ \boldsymbol X_2^T \boldsymbol X_2 \in R^{m \times m}$ are actually constant,  which require  $2m^2l$ flam and $2m^2N_2$ flam.  
Then, the optimal $\boldsymbol P_{{\mathcal M}_2}$ is acquired after $t$ times iteration.
The calculation of $\boldsymbol R_k$ needs $2m^2l+ml$ flam and the SVD of $\boldsymbol R_k$ requires $\dfrac{3}{2}l^2m+\dfrac{9}{2}l^3$ flam.
The calculation of matrix $\boldsymbol W_k \boldsymbol I_{m,l} \boldsymbol V_k^T$ requires $2m^2l$ flam. Overall, the time complexity of Algorithm \ref{ALG1} is $(4m^2l+\dfrac{3}{2}l^2m+\dfrac{9}{2}l^3+ml)t+2m^2l+2m^2N_2$  flam.
PCA-EWC adds at most $(4m^2l+\dfrac{3}{2}l^2m+\dfrac{9}{2}l^3+ml)t+2m^2l$ flam compared with PCA.
In practical applications, the initial setting of Algorithm \ref{ALG1} makes it converge fast.  

\subsubsection{Potential limitation and solution}

This method requires the existence of similarity in different modes, thus the retained information about previous modes would be useful for new modes. This requirement is relatively easy to satisfy for industrial systems, because the systems follow similar physical or chemical laws for different operating modes. Obviously, the performance of PCA-EWC would be affected by similarity in operating modes. The performance may keep excellent if data from various modes are similar. However, if data from new mode are completely different from the previous modes, this method can not deliver excellent performance. Aimed at this case, the monitoring model would be retrained  or the transfer learning  would be adopted if inner relationship between the new mode and the previous ones can be found.

Note that the order of operating modes also influences the monitoring performance to a certain degree, because the proposed method is more reliable for recent modes than earlier ones. But for neighboring modes, the mode order has considerably limited effect for performance because the extracted information is similar and sufficient.

\section{Experimental study} \label{Sec:4}
This section adopts a numerical case and a practical pulverizing system to illustrate the continual learning ability of  PCA-EWC for monitoring multimode processes. Through  different combinations of training data and testing data in Table \ref{Table-simulation}, the continual learning ability of PCA-EWC is illustrated. Besides, recursive PCA (RPCA), Gaussian mixture models (GMMs) and improved mixture of probabilistic principal component analysis (IMPPCA) \cite{zhang2019an} are adopted to compare with the proposed method.

\subsection{Numerical case study}\label{case:numerical}

We employ the following case \cite{tong2017a}:

\begin{equation}
	\left[ {\begin{array}{*{20}{c}}
			{{x_1}}\\
			{{x_2}}\\
			{{x_3}}\\
			{{x_4}}\\
			{{x_5}}\\
			{{x_6}}\\
			{{x_7}}\\
			{{x_8}}
	\end{array}} \right] = \left[ {\begin{array}{*{20}{c}}
			{0.95}&{0.82}&{0.94}\\
			{0.23}&{0.45}&{0.92}\\
			{ - 0.61}&{0.62}&{0.41}\\
			{0.49}&{0.79}&{0.89}\\
			{0.89}&{ - 0.92}&{0.06}\\
			{0.76}&{0.74}&{0.35}\\
			{0.46}&{0.58}&{0.81}\\
			{ - 0.02}&{0.41}&{0.01}
	\end{array}} \right]\left[ {\begin{array}{*{20}{c}}
			{{s_1}}\\
			{{s_2}}\\
			{{s_3}}
	\end{array}} \right] + \boldsymbol e
\end{equation}
where the  noise $\boldsymbol e \in R^8$ follows Gaussian distribution  $ e_i \sim \mathbb{N}(0,0.001), i=1,\cdots,8$.  We generate four sequential data successively as follows: \\
$\bullet$   Data 1: $s_1 \sim \mathbb{U}([-10,-9.7])$, $ s_2 \sim \mathbb{N}(-5,1)$, and $s_3 \sim \mathbb{U}([2,3])$; 	\\
$\bullet$   Data 2: $s_1 \sim \mathbb{U}([-6,-5.7])$, $ s_2 \sim \mathbb{N}(-1,1)$, and $s_3 \sim \mathbb{U}([3,4])$;  	\\
$\bullet$   Data 3: $s_1 \sim \mathbb{U}([-10,-9.7])$, $ s_2 \sim \mathbb{N}(-5,1)$, and $s_3 \sim \mathbb{U}([2,3])$;\\
$\bullet$   Data 4: $s_1 \sim \mathbb{U}([-9,-8.7])$, $ s_2 \sim \mathbb{N}(-5,1)$, and $s_3 \sim \mathbb{U}([3,4])$.

\noindent where $\mathbb{U}([-10,-9.7])$ represents the uniform distribution between $-10$ and $-9.7$, and so on.
Data 1 and 3 follow the same distribution and come from the mode ${\mathcal M}_1$.  Data 2 are collected from the mode ${\mathcal M}_2$. Obviously, Data 4 have a certain degree of similarity with Data 1 and 2, which can also be verified and measured by Wasserstein distance.

1000 normal samples from Data 1 and 2 are generated to train the model, denoted as training data 1 and  2, respectively. 1000 samples from Data $i$, $i=1,2,3,4$, are collected for testing and the novelty scenarios are designed below: \\
$\bullet$  Fault 1:   $x_3$  is added by $0.1$ from $501th$ sample;\\
$\bullet$  Fault 2:   $x_6$  is added by $0.1$ from $501th$ sample;\\
$\bullet$  Fault 3:  the slope drift occurs in  $x_1$ from $501th$ sample and the slope rate is 0.002.

Take  Fault 1 for instance, the variable $x_3$ is added by 0.1 in Data $i$, $i=1,2,3,4$, which constitutes the testing data $i$. Four situations  are designed to illustrate the effectiveness of PCA-EWC, as depicted in Table \ref{Table-simulation}.
Here, we define Situation 5
to illustrate the effectiveness of PCA-EWC on similar modes.   RPCA, GMMs and IMPPCA are adopted to compare with the proposed method, as designed in Table \ref{Table1-comparative}.

\begin{table}[!hbp]
	\begin{center}
		\caption{Comparative scheme for numerical case}\label{Table1-comparative}
		\footnotesize
		\begin{tabular}{c c c c c}
			\hline
			& Methods &  \makecell{Training data \\sources}   & \makecell{Testing data \\sources}  \\
			\hline
			Situation 5 & PCA-EWC & Data 2 and Model A & Data 4\\
			Situation 6 & RPCA & Data 1 and  Data 2    & Data 3  \\
			Situation 7 & RPCA & Data 2 and  Data 3   & Data 4 \\
			Situation 8 & GMMs & Data 1 and  Data 2    & Data 3    \\
			Situation 9 & GMMs & Data 1 and  Data 2   & Data 4    \\
			Situation 10 & IMPPCA & Data 1 and  Data 2    & Data 3    \\
			Situation 11 & IMPPCA & Data 1 and  Data 2   & Data 4    \\			
			\hline
		\end{tabular}
	\end{center}
	\vspace{-0.5cm}
\end{table}

\begin{table*}[!htbp]
	\begin{center}
		\caption{Evaluation indexes of the numerical case study}\label{Table1}
		\footnotesize
		\begin{tabular}{c c c c c c c c c c c } 
			\hline
			\multirow{1}{*}{Fault type} & \multirow{2}{*} {\makecell{Statistic \\index}}  &\multicolumn{3}{c}{Fault 1}  & \multicolumn{3}{c}{Fault 2}  & \multicolumn{3}{c}{Fault 3}    \\
			\cmidrule(r){1-1}    \cmidrule(r){3-5}  \cmidrule(r){6-8} \cmidrule(r){9-11}
			Indexes &  &\multicolumn{1}{c} {FDR}    & {FAR}   & \multicolumn{1}{c}{DD}   &\multicolumn{1}{c} {FDR}   &FAR   & \multicolumn{1}{c}{DD}   & \multicolumn{1}{c} {FDR}    &FAR   & \multicolumn{1}{c}{DD}   \\  
			\cline{1-11}
			\multirow{2}{*}{Situation 1} & mean & \multicolumn{1}{c} {100}  & $<10^{-2}$ & \multicolumn{1}{c}{0}
			& \multicolumn{1}{c} {100}  & $<10^{-2}$ & \multicolumn{1}{c}{0}
			& \multicolumn{1}{c} {98.82}  & $<10^{-2}$& \multicolumn{1}{c}{5.8210} \\
			 & std & \multicolumn{1}{c} {0}  & $<10^{-2}$ & \multicolumn{1}{c}{0}
			 & \multicolumn{1}{c} {0}  & $<10^{-2}$ & \multicolumn{1}{c}{0}
			 & \multicolumn{1}{c} {0.14}  & $<10^{-2}$& \multicolumn{1}{c}{0.7639} \\
			%
		\multirow{2}{*}{Situation 2}	& mean& \multicolumn{1}{c} {100}  & 0 & \multicolumn{1}{c}{0}
			&\multicolumn{1}{c}  {100}  & $<10^{-2}$ & \multicolumn{1}{c}{0}
			& \multicolumn{1}{c} {95.52}  & 0 & \multicolumn{1}{c}{{5.8210}} \\
			          &std & \multicolumn{1}{c} {0}  & 0 & \multicolumn{1}{c}{0}
			&\multicolumn{1}{c}  {0}  & $<10^{-2}$ & \multicolumn{1}{c}{0}
			& \multicolumn{1}{c} {0.27}  & 0 & \multicolumn{1}{c}{{0.7639}} \\
			%
		\multirow{2}{*}	{Situation 3} & mean& \multicolumn{1}{c} {100}  &2.47 & \multicolumn{1}{c}{0}
			& \multicolumn{1}{c} {100}  &2.67 & \multicolumn{1}{c}{0}
			& \multicolumn{1}{c} {95.90}  & 2.57& \multicolumn{1}{c}{17.74} \\
				 & std & \multicolumn{1}{c} {0}  &6.81 & \multicolumn{1}{c}{0}
			& \multicolumn{1}{c} {0}  &6.96 & \multicolumn{1}{c}{0}
			& \multicolumn{1}{c} {0.77}  & 6.91& \multicolumn{1}{c}{6.7710} \\		
			\multirow{2}{*}	{Situation 4}& mean& \multicolumn{1}{c} {100}  &44.39 & \multicolumn{1}{c}{0}
			& \multicolumn{1}{c} {100}   & 45.38 & \multicolumn{1}{c}{0}
			& \multicolumn{1}{c} {99.44}  &44.44& \multicolumn{1}{c}{17.74} \\
		   	& std& \multicolumn{1}{c} {0}  &25.83 & \multicolumn{1}{c}{0}
			& \multicolumn{1}{c} {0}   & 25.55 & \multicolumn{1}{c}{0}
			& \multicolumn{1}{c} {0.37}  &25.59& \multicolumn{1}{c}{6.7710} \\
			%
			\multirow{2}{*}	{Situation 5} & mean& \multicolumn{1}{c} {100}  &2.47 & \multicolumn{1}{c}{0}
			& \multicolumn{1}{c} {100}  &2.68 & \multicolumn{1}{c}{0}
			& \multicolumn{1}{c} {95.91}  & 2.56& \multicolumn{1}{c}{17.7423} \\
			& std& \multicolumn{1}{c} {0}  &6.81 & \multicolumn{1}{c}{0}
			& \multicolumn{1}{c} {0}  &6.96 & \multicolumn{1}{c}{0}
			& \multicolumn{1}{c} {0.74}  & 6.90& \multicolumn{1}{c}{6.7680} \\
			%
			\multirow{2}{*}	{Situation 6} & mean & \multicolumn{1}{c} {100}  &100 & \multicolumn{1}{c}{0}
			& \multicolumn{1}{c} {100}  &100 & \multicolumn{1}{c}{0}
			& \multicolumn{1}{c} {100}  &100& \multicolumn{1}{c}{0} \\
			& std & \multicolumn{1}{c} {0}  &$<10^{-2}$ & \multicolumn{1}{c}{0}
			& \multicolumn{1}{c} {0}  &0 & \multicolumn{1}{c}{0}
			& \multicolumn{1}{c} {0}  &0& \multicolumn{1}{c}{0} \\
			%
			\multirow{2}{*}	{Situation 7} & mean&\multicolumn{1}{c} {100}  &100 & \multicolumn{1}{c}{0}
			& \multicolumn{1}{c} {100}  &100 & \multicolumn{1}{c}{0}
			& \multicolumn{1}{c}{100}  &100& \multicolumn{1}{c}{0} \\
			& std&\multicolumn{1}{c} {0}  &0 & \multicolumn{1}{c}{0}
			& \multicolumn{1}{c} {0}  &0 & \multicolumn{1}{c}{0}
			& \multicolumn{1}{c}{0}  &0& \multicolumn{1}{c}{0} \\
			\multirow{2}{*}	{Situation 8} & mean&\multicolumn{1}{c} {100}  &22.68 & \multicolumn{1}{c}{0}
			& \multicolumn{1}{c} {100}  &23.92 & \multicolumn{1}{c}{0}
			& \multicolumn{1}{c} {97.69}  & 23.80 & \multicolumn{1}{c}{9.5260} \\
			& std&\multicolumn{1}{c} {0}  &27.98 & \multicolumn{1}{c}{0}
			& \multicolumn{1}{c} {0}  &28.11 & \multicolumn{1}{c}{0}
			& \multicolumn{1}{c} {1.86}  & 28.25& \multicolumn{1}{c}{9.3606} \\
			%
			\multirow{2}{*}	{Situation 9} & mean&\multicolumn{1}{c} {99.27}  &75.38 & \multicolumn{1}{c}{0.120}
			& \multicolumn{1}{c} {100}  &100 & \multicolumn{1}{c}{0}
			& \multicolumn{1}{c} {99.16}  & 77.22   & \multicolumn{1}{c}{2.8720} \\
			 & std&\multicolumn{1}{c} {5.32}  &38.30 & \multicolumn{1}{c}{0.1895}
			& \multicolumn{1}{c} {0}  &0 & \multicolumn{1}{c}{0}
			& \multicolumn{1}{c} {1.58}  & 37.55   & \multicolumn{1}{c}{6.8828} \\
			\multirow{2}{*}	{Situation 10} & mean&\multicolumn{1}{c} {97.23}  &1.83 & \multicolumn{1}{c}{1.5670}
			& \multicolumn{1}{c} {96.21}  &1.87 & \multicolumn{1}{c}{1.1610}
			& \multicolumn{1}{c} {96.95}  & 1.83& \multicolumn{1}{c}{3.53} \\
			 & std &\multicolumn{1}{c} {15.68}  &0.82 & \multicolumn{1}{c}{13.8753}
			& \multicolumn{1}{c} {17.95}  &0.85 & \multicolumn{1}{c}{9.0722}
			& \multicolumn{1}{c} {13.20}  &0.83& \multicolumn{1}{c}{10.6229} \\
			%
			\multirow{2}{*}	{Situation 11} &mean &\multicolumn{1}{c} {98.70}  &50.99 & \multicolumn{1}{c}{0.13}
			& \multicolumn{1}{c} {97.94}  &50.21 & \multicolumn{1}{c}{0.7430}
			& \multicolumn{1}{c} {98.47}  &51.20   & \multicolumn{1}{c}{0.7940} \\
			 &std &\multicolumn{1}{c} {8.03}  &10.44 & \multicolumn{1}{c}{3.3913}
			& \multicolumn{1}{c} {10.92}  &9.89 & \multicolumn{1}{c}{15.6185}
			& \multicolumn{1}{c} {8.09}  &11.06   & \multicolumn{1}{c}{0.9862} \\
			\cline{1-11}
		\end{tabular}
	\end{center}
\end{table*}

\begin{figure*}[!tbp]
	\centering
	\subfigure{\label{fault1-1}}\addtocounter{subfigure}{-2}
	\subfigure
	{\subfigure[Situation 1]{\includegraphics[width=0.235\textwidth]{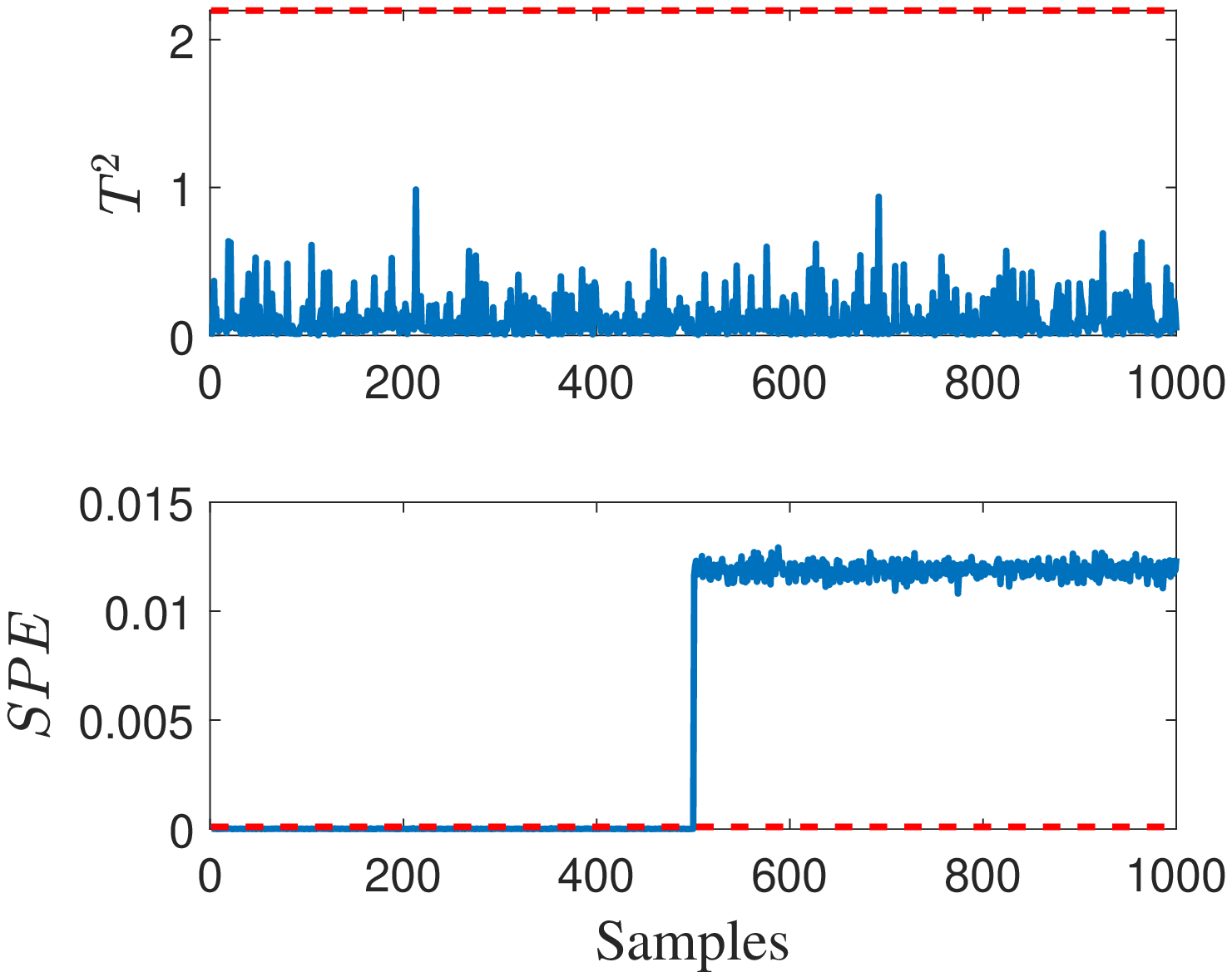}}}
	\vspace{-1.5mm}
	\hspace{-1mm}
	\subfigure{\label{fault1-2}}\addtocounter{subfigure}{-2}
	\subfigure
	{\subfigure[Situation 2]{\includegraphics[width=0.235\textwidth]{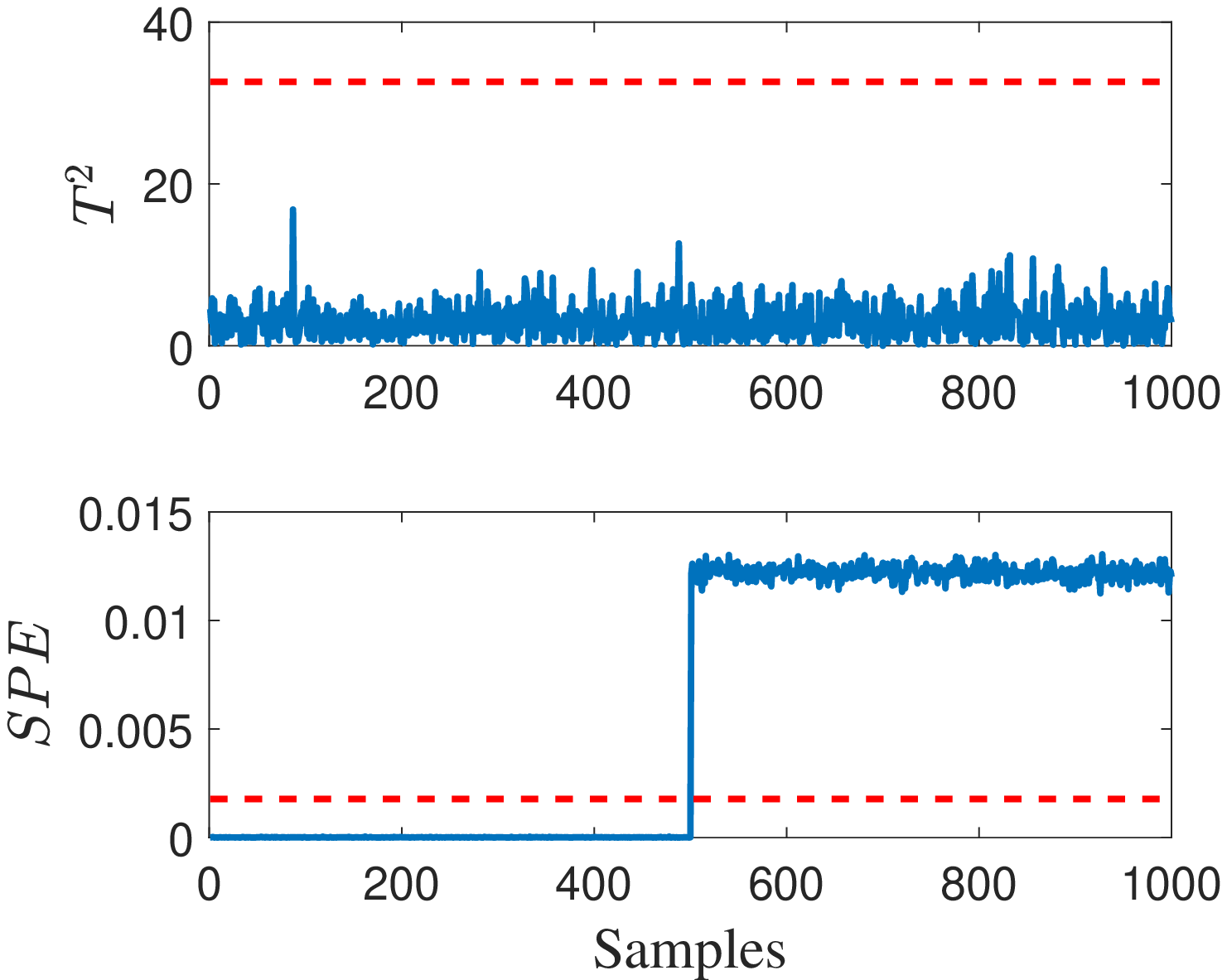}}}
	%
	\hspace{-1mm}
	\vspace{-1.5mm}
	\subfigure{\label{fault1-3}}\addtocounter{subfigure}{-2}
	\subfigure
	{\subfigure[Situation 3]{\includegraphics[width=0.235\textwidth]{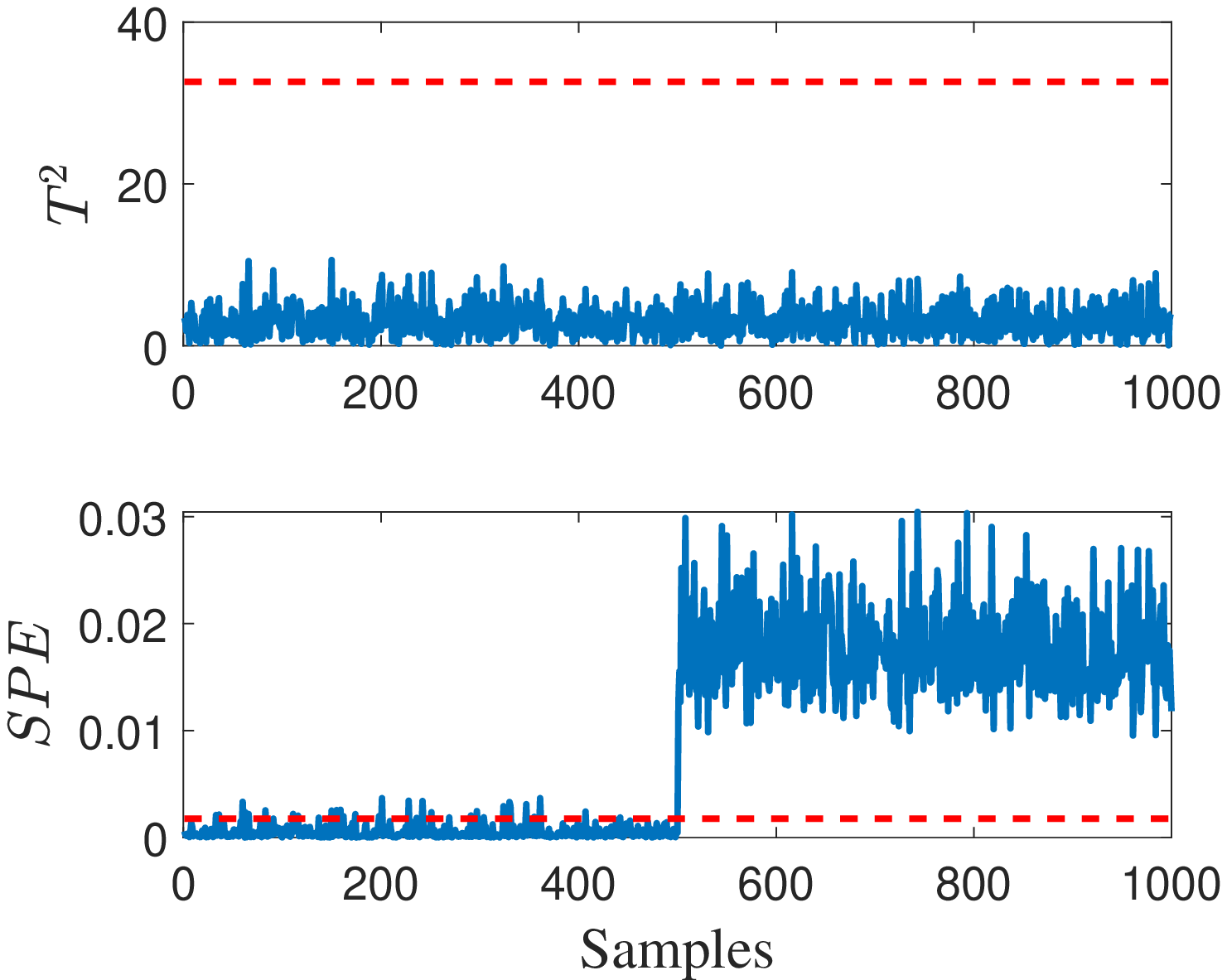}}}
	\subfigure{\label{fault1-4}}\addtocounter{subfigure}{-2}
	\hspace{-1mm}
	\vspace{-1.5mm}
	\subfigure
	{\subfigure[Situation 4]{\includegraphics[width=0.235\textwidth]{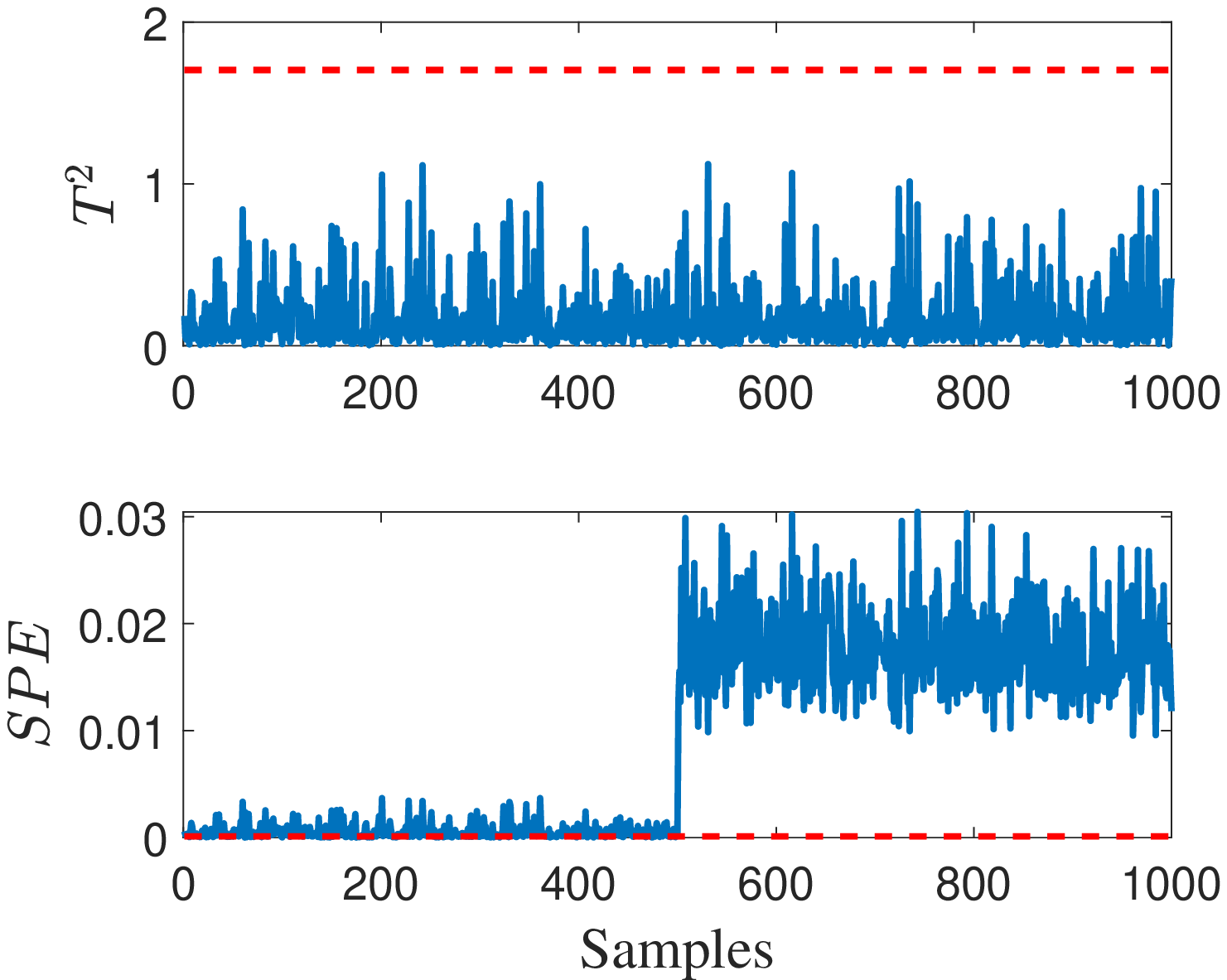}}}
	\hspace{-1mm}
	\vspace{-1.5mm}
	\subfigure{\label{fault1-5}}\addtocounter{subfigure}{-2}
	\subfigure
	{\subfigure[Situation 5]{\includegraphics[width=0.235\textwidth]{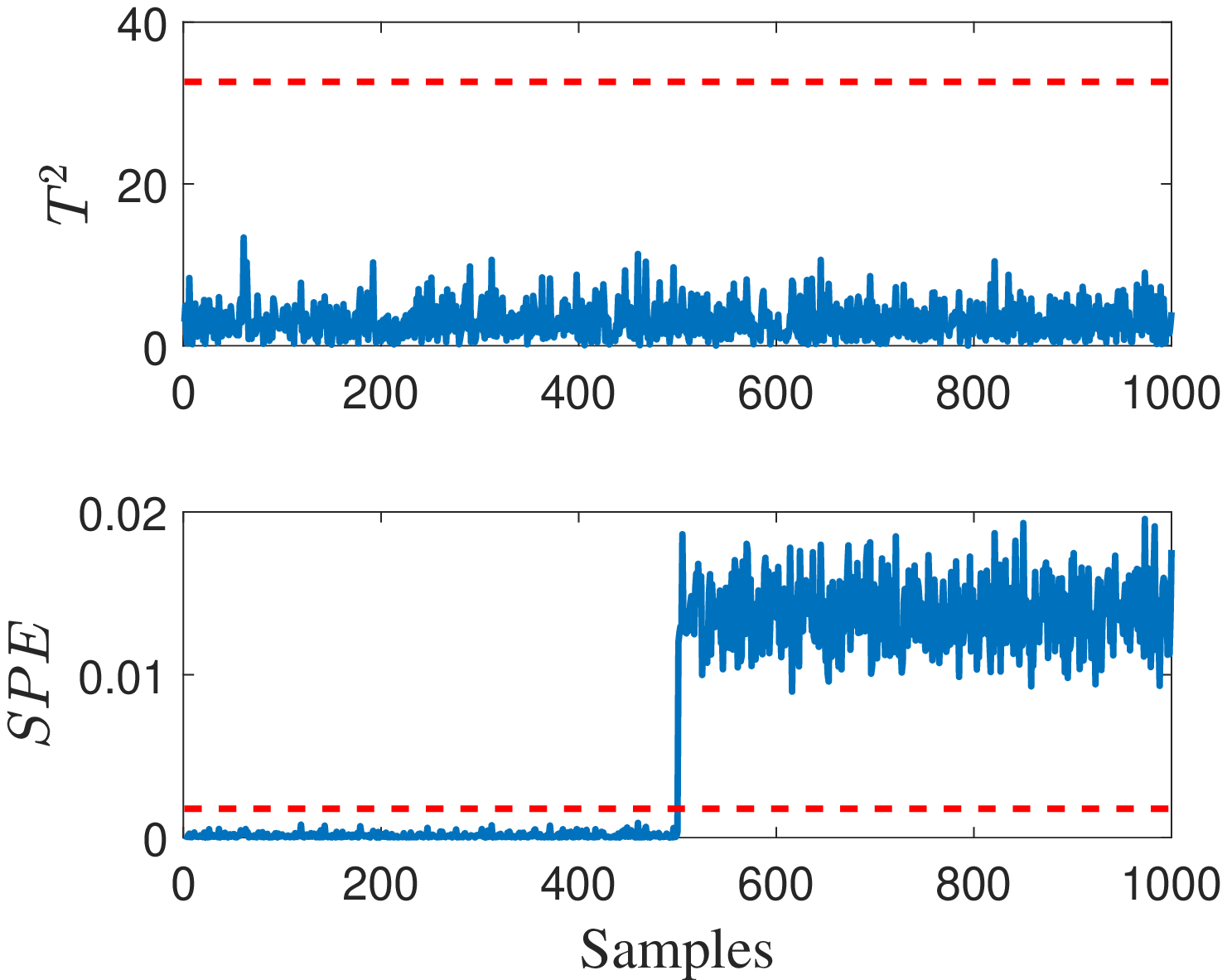}}}
	\subfigure{\label{fault1-6}}\addtocounter{subfigure}{-2}
	\subfigure
	{\subfigure[Situation 6]{\includegraphics[width=0.235\textwidth]{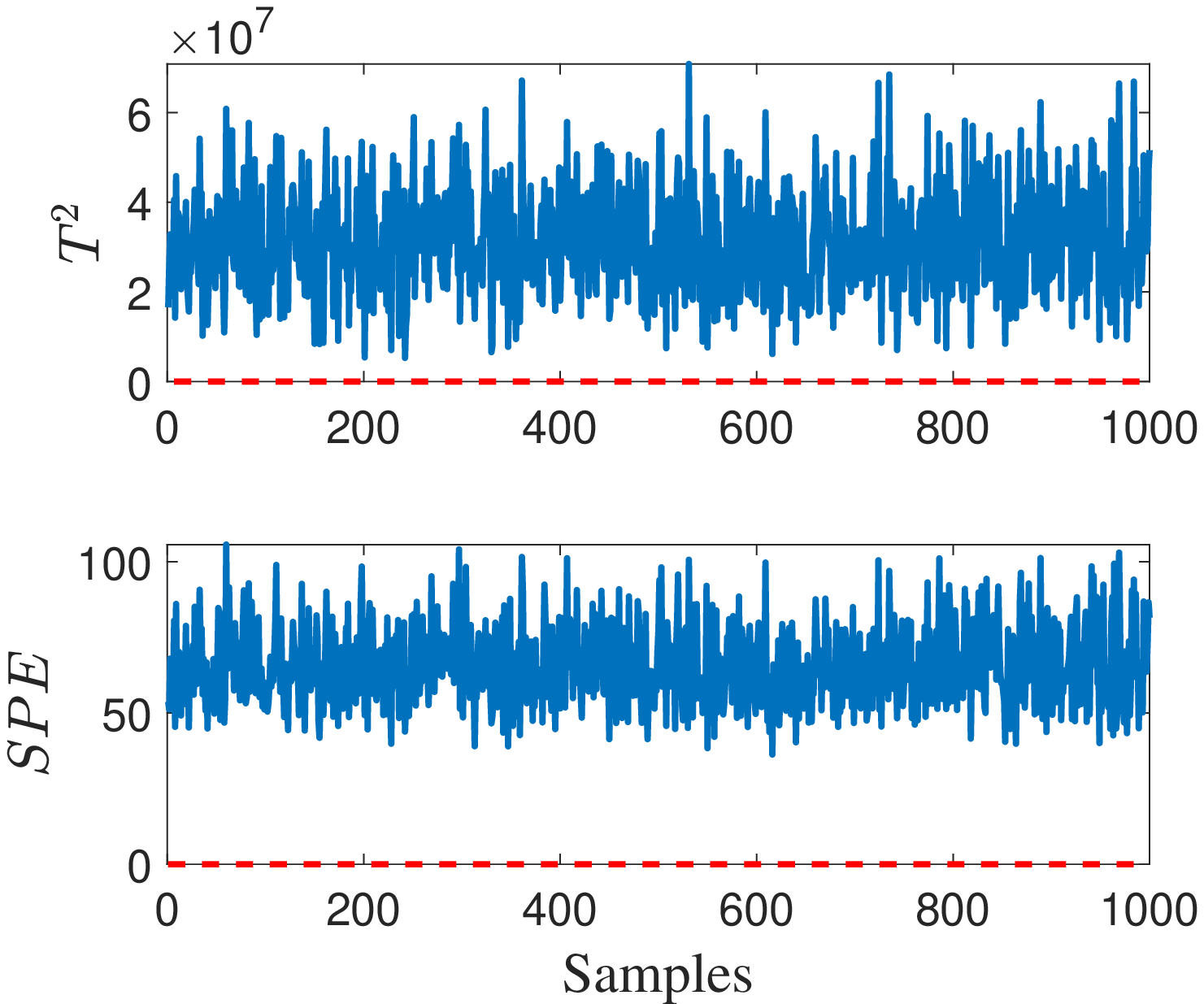}}}
	\hspace{-1mm}
	\vspace{-1.5mm}
	\subfigure{\label{fault1-7}}\addtocounter{subfigure}{-2}
	\subfigure
	{\subfigure[Situation 7]{\includegraphics[width=0.235\textwidth]{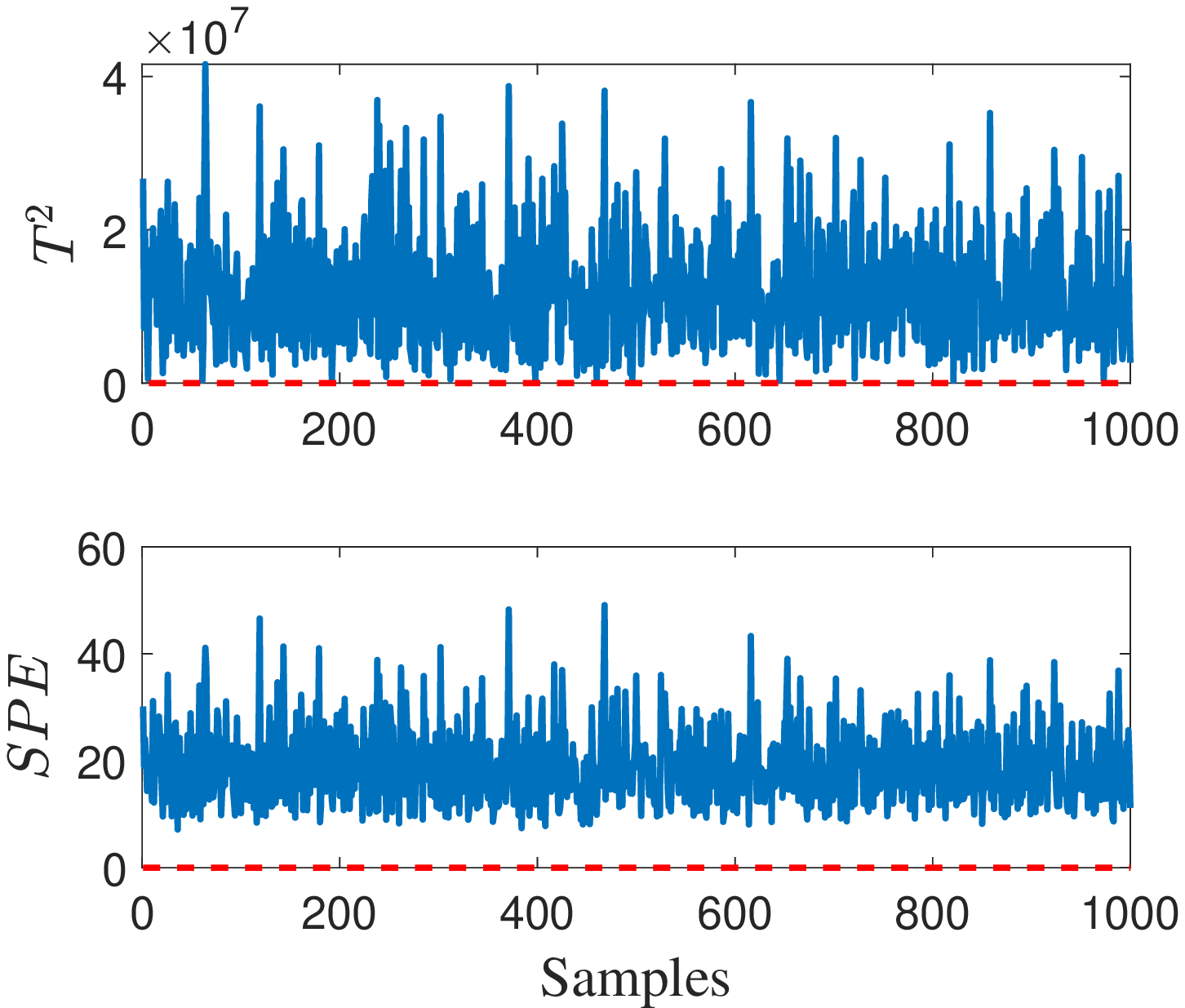}}}
	\subfigure{\label{fault1-8}}\addtocounter{subfigure}{-2}
	\subfigure
	{\subfigure[Situation 8]{\includegraphics[width=0.235\textwidth]{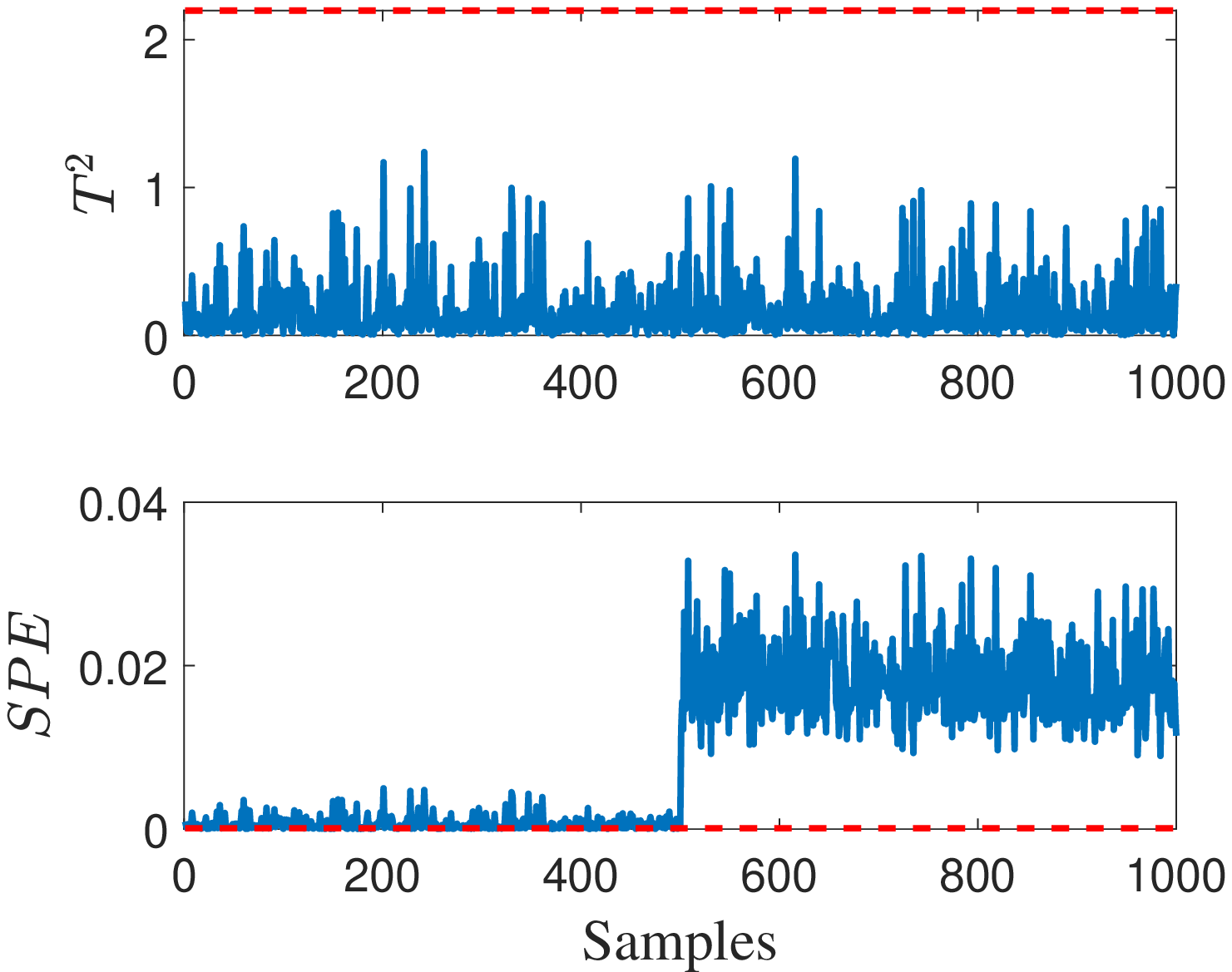}}}
	\hspace{-1mm}
	\vspace{-1.5mm}
	\subfigure{\label{fault1-9}}\addtocounter{subfigure}{-2}
	\subfigure
	{\subfigure[Situation 9]{\includegraphics[width=0.235\textwidth]{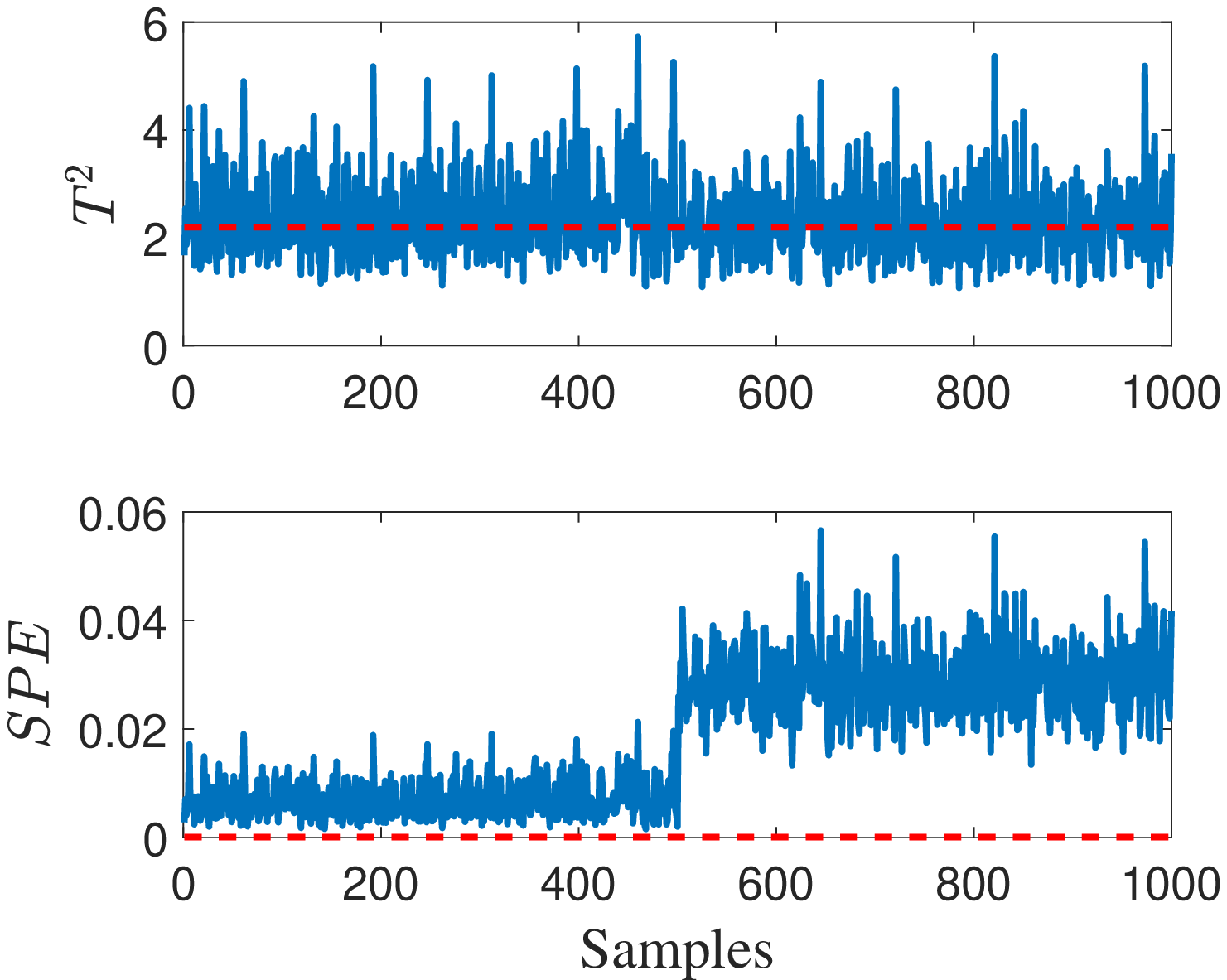}}}
	\hspace{-1mm}
	\vspace{-1.5mm}
	\subfigure{\label{fault1-10}}\addtocounter{subfigure}{-2}
	\subfigure
	{\subfigure[Situation 10]{\includegraphics[width=0.235\textwidth]{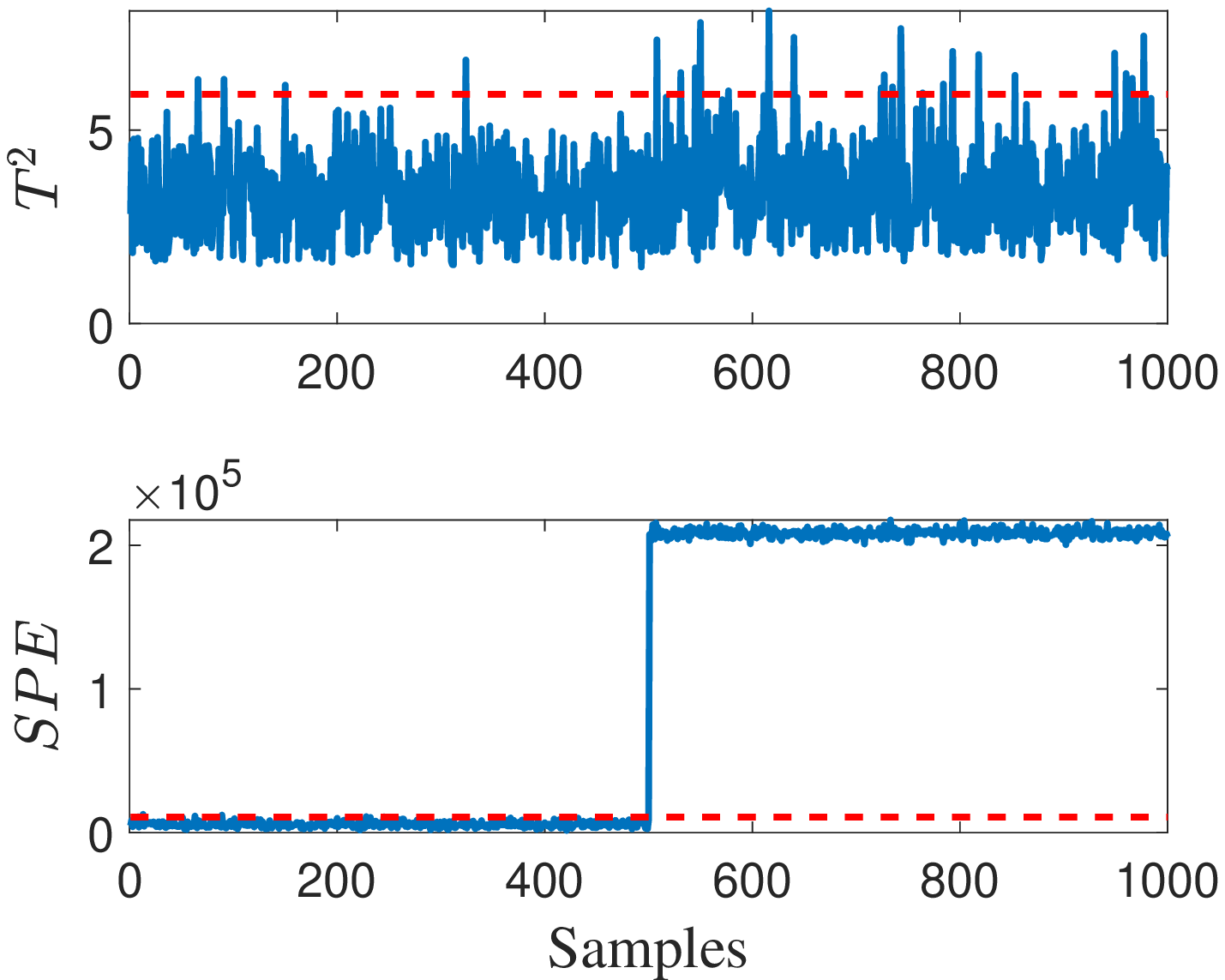}}}
	\hspace{-1mm}
	\vspace{-1.5mm}
	\subfigure{\label{fault1-11}}\addtocounter{subfigure}{-2}
	\subfigure
	{\subfigure[Situation 11]{\includegraphics[width=0.238\textwidth]{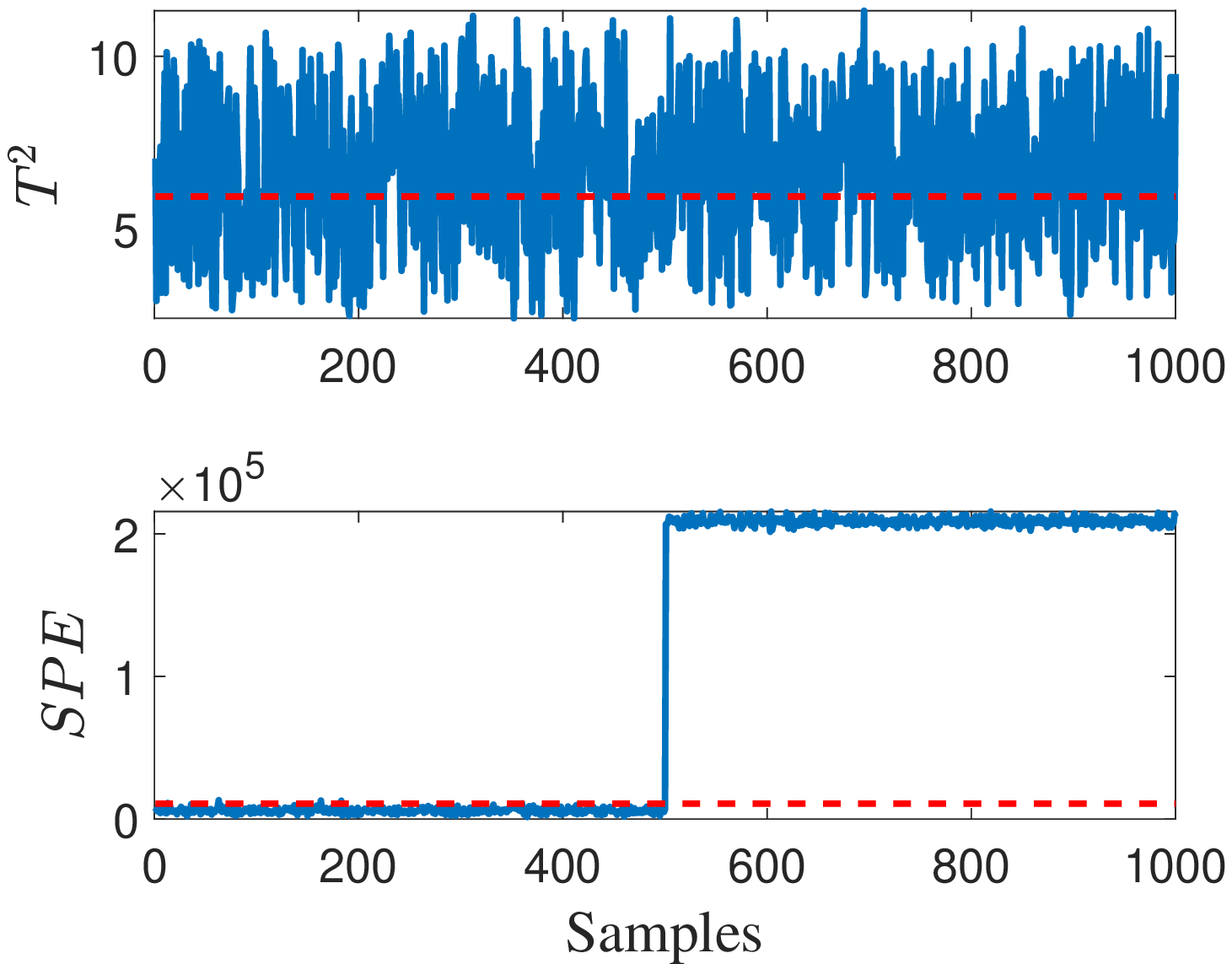}}}
	\centering
	\caption{Monitoring charts of Fault 1} \label{case1}
\end{figure*}

Without loss of generality, we conduct 1000 independent  repeated experiments to enhance reliability.
The results, including mean value and standard deviation (std) of FDR($\%$),  FAR($\%$) and DD, are calculated in Table \ref{Table1}. The consequences of Fault 1 and Fault 2 are similar. Besides, faults can be detected accurately and timely, expect for Situation 4. For Fault 3, the FDRs of Situations 1-3 and 5 for Fault 3 are less than $100\%$ because the abnormal amplitude at initial abnormal time is so small that can not be detected immediately. However, the detection effect of Situation 4 is still poor when the abnormal amplitude increases. For Situations 6-7, the FARs are pretty high and even approach to $100\%$, which indicate that RPCA  can not separate fault from normal change and track the system. 
 For GMMs, the FAR of previous trained mode is over $20\%$. Meanwhile, the FAR of similar mode is higher than $70\%$, which indicates that GMMs fails to monitor the same or similar modes accurately.
For Situations 10-11, IMPPCA can monitor the trained mode accurately but fails to detect the fault in a similar mode.

Then, we select one independent experiment to explain the performance specifically.
Take the Fault 1 as an instance to explain the performance specifically and the partial simulation results are illustrated in Figure \ref{case1}.
The FDR of Situation 1 is $100\%$  and it indicates that we get an accurate training model for the mode ${\mathcal M}_1$. Outstanding performance is also reflected on Situation 2 in Figure \ref{fault1-2}, which implies that PCA-EWC is effective for monitoring the mode ${\mathcal M}_2$.  The Model B performs excellently on the mode ${\mathcal M}_1$ and the FDR of Situation 3 is $100\%$.
It signifies that partial information of the mode ${\mathcal M}_1$ is retained when PCA is utilized for the current mode ${\mathcal M}_2$, which is sufficient to provide optimal performance.  Combining Situations 2 and 3, we discover that the Model B is effective for monitoring modes ${\mathcal M}_1$ and ${\mathcal M}_2$ simultaneously. However, the performance of Situation 4 is especially poor and the FAR is higher than $44\%$. Comparing Situation 4 with Situation 3, the learned knowledge from the mode ${\mathcal M}_1$ has been forgotten visually and the performance decreases disastrously by PCA. The monitoring model by PCA in one mode fails to detect novelty in another mode. The result of Situation 5  is pretty excellent, which illustrates that PCA-EWC  provides optimal performance and continual learning ability for similar modes, as illustrated in Figure \ref{fault1-5}.

RPCA and GMMs are not able to distinguish faults and normal data  in Figures \ref{fault1-6}-\ref{fault1-9}.   For Situations 6, although RPCA has trained the mode ${\mathcal M}_1$, it still fails to monitor the same mode because the information is overwritten by the mode ${\mathcal M}_2$. For Situation 7, the result illustrates that RPCA is not capable of monitoring the similar mode. That is to say, RPCA is not able to track the mode change quickly. GMMs forgets the information of previous modes and fails to monitor the same or similar modes. IMPPCA  is able to monitor the previous trained mode, as shown in Figure \ref{fault1-10}.  However, it fails to monitor the similar mode that is not trained before and the FAR is over $50\%$.
Generally speaking, RPCA, GMMs and IMPPCA suffer from the ``catastrophic forgetting'' issue.
The analysis aforementioned is also applicable to Fault 2 and Fault 3. There exists detection delay for Fault 3 and  the expected means are 5.82 or 17.8, which are acceptable because the fault amplitudes are 0.012 and 0.035, respectively.

According to the above-mentioned analysis, PCA-EWC can preserve significant information from the previous monitoring tasks when learning the new monitoring task, thus catastrophic forgetting of PCA is overcome for successive modes. Besides, the retained partial information is adequate to deliver optimal performance and thus PCA-EWC is capable of monitoring the same or similar modes. In brief, PCA-EWC can achieve favorable capability for monitoring successive modes, without retraining from scratch.

\subsection{Pulverizing system process monitoring}
\begin{figure}[!tbp]
	\centering
	\includegraphics[width=0.28\textwidth]{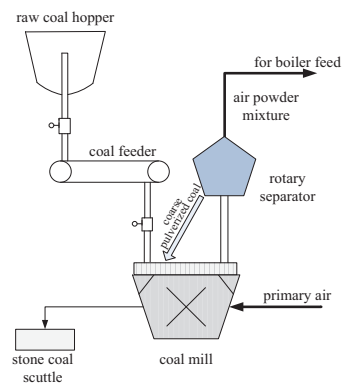}
	\caption{Schematic diagram of the coal pulverizing system}
	\label{fig_benchmark}
\end{figure}

The 1000-MW ultra-supercritical thermal power plant is increasingly popular and highly complex. In this paper, we investigate one important unit of boiler, namely, the coal pulverizing system. The coal pulverizing system in Zhoushan Power Plant, Zhejiang Province, China, includes coal feeder, coal mill, rotary separator, raw coal hopper and stone coal scuttle, as depicted in Figure \ref{fig_benchmark}. It is expected to provide the proper pulverized coal with desired coal fineness and optimal temperature. The operating modes would vary  owing to the types of coal and change of unit load.

\begin{table*}[!tbp]
	
	\begin{center}
		\caption{Fault information and experimental data of the practical coal pulverizing system}\label{Table_informationZS}
		\footnotesize
		\renewcommand\arraystretch{1.25}
		\begin{tabular}{c| c| c c c c} 
			\hline
			\makecell{Fault\\ type} & Key variables &  \makecell{Mode\\ number} & \makecell{NoTrS/NoTeS} &\makecell{Fault\\ location}  &Fault cause  \\
			\hline
			\multirow{3}{*} {\makecell{Fault \\4}} &\multirow{3}{4.5cm}{Nine variables: outlet temperature, pressure of air powder mixture,  primary air temperature, hot/cold primary air  pressure, etc.}
			& \makecell{${\mathcal M}_1$} &2160/2880 & 909  & \makecell{Internal deflagration  due to high temperature} \\
			& & \makecell{ ${\mathcal M}_2$} & 1080/1080 & 533 & \makecell{Hot primary air electric damper failure}   \\ 
			&  & \makecell{ ${\mathcal M}_3$}  &0/1440 & 626   &\makecell{Air leakage at primary air interface}\\
			\hline		
			\multirow{3}{*}{\makecell{Fault\\ 5}} &\multirow{3}{4.5cm} {Nine variables: rotary separator speed and current, coal feeding capacity, bearing temperature, etc.}
			&\makecell{ ${\mathcal M}_1$}  & 2880/1080 & 806 &  \makecell{Frequency conversion cabinet  short circuit} \\
			&   & \makecell{ ${\mathcal M}_2$}  & 720/720 & 352 & \makecell{High temperature of separator bearing} \\ 
			&   & \makecell{ ${\mathcal M}_3$}  & 0/2160 & 134 &  Large vibration  \\
			\hline
		\end{tabular}
	\end{center}
\end{table*}

\begin{table}[!htp]
	\begin{center}
		\caption{Comparative scheme for the coal pulverizing system}\label{Table2-comparative}
		\footnotesize
		\begin{tabular}{c c c c c}
			\hline
			& Methods &  \makecell{Training data \\sources}   & \makecell{Testing data \\sources}  \\
			\hline
			Situation 5 & RPCA & Mode ${\mathcal M}_1$         & Mode ${\mathcal M}_2$         \\
			Situation 6 & RPCA & Modes ${\mathcal M}_1$ and ${\mathcal M}_2$  & Mode ${\mathcal M}_3$ \\
			Situation 7 & GMMs & Modes ${\mathcal M}_1$ and ${\mathcal M}_2$  & Mode ${\mathcal M}_2$      \\
			Situation 8 & GMMs & Modes ${\mathcal M}_1$ and ${\mathcal M}_2$  & Mode ${\mathcal M}_3$     \\
			Situation 9 & IMPPCA & Modes ${\mathcal M}_1$ and ${\mathcal M}_2$  & Mode ${\mathcal M}_2$      \\
            Situation 10 & IMPPCA & Modes ${\mathcal M}_1$ and ${\mathcal M}_2$  & Mode ${\mathcal M}_3$     \\
			\hline
		\end{tabular}
	\end{center}
\end{table}

    \begin{figure*}[!hbp]
    	\centering
    	\subfigure{\label{variable-1}}\addtocounter{subfigure}{-2}
    	\subfigure
    	{\subfigure[Primary air temperature]{\includegraphics[width=0.230\textwidth]{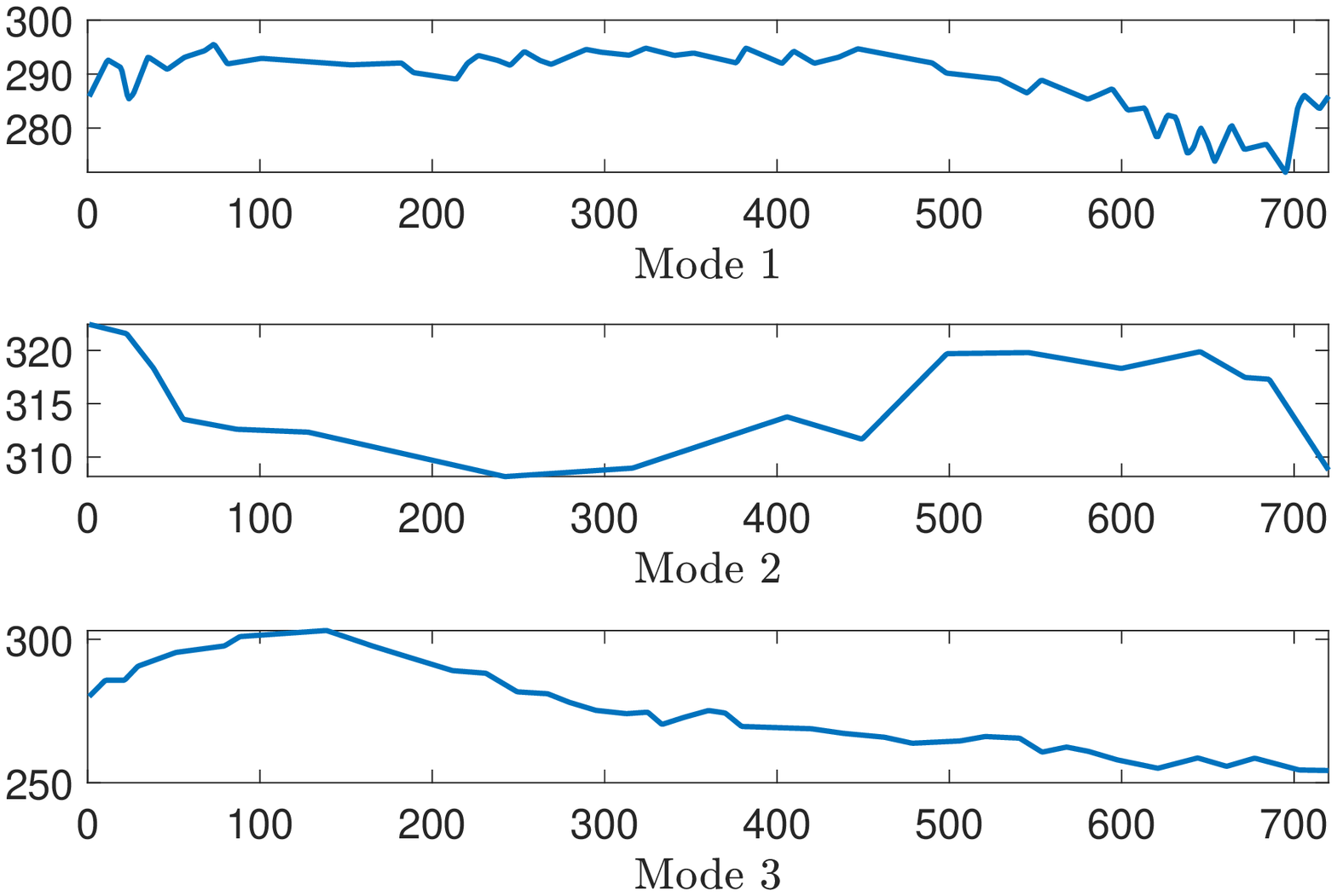}}}
    	\vspace{-1mm}
    	\subfigure{\label{variable-2}}\addtocounter{subfigure}{-2}
    	\subfigure
    	{\subfigure[Primary air pressure]{\includegraphics[width=0.235\textwidth]{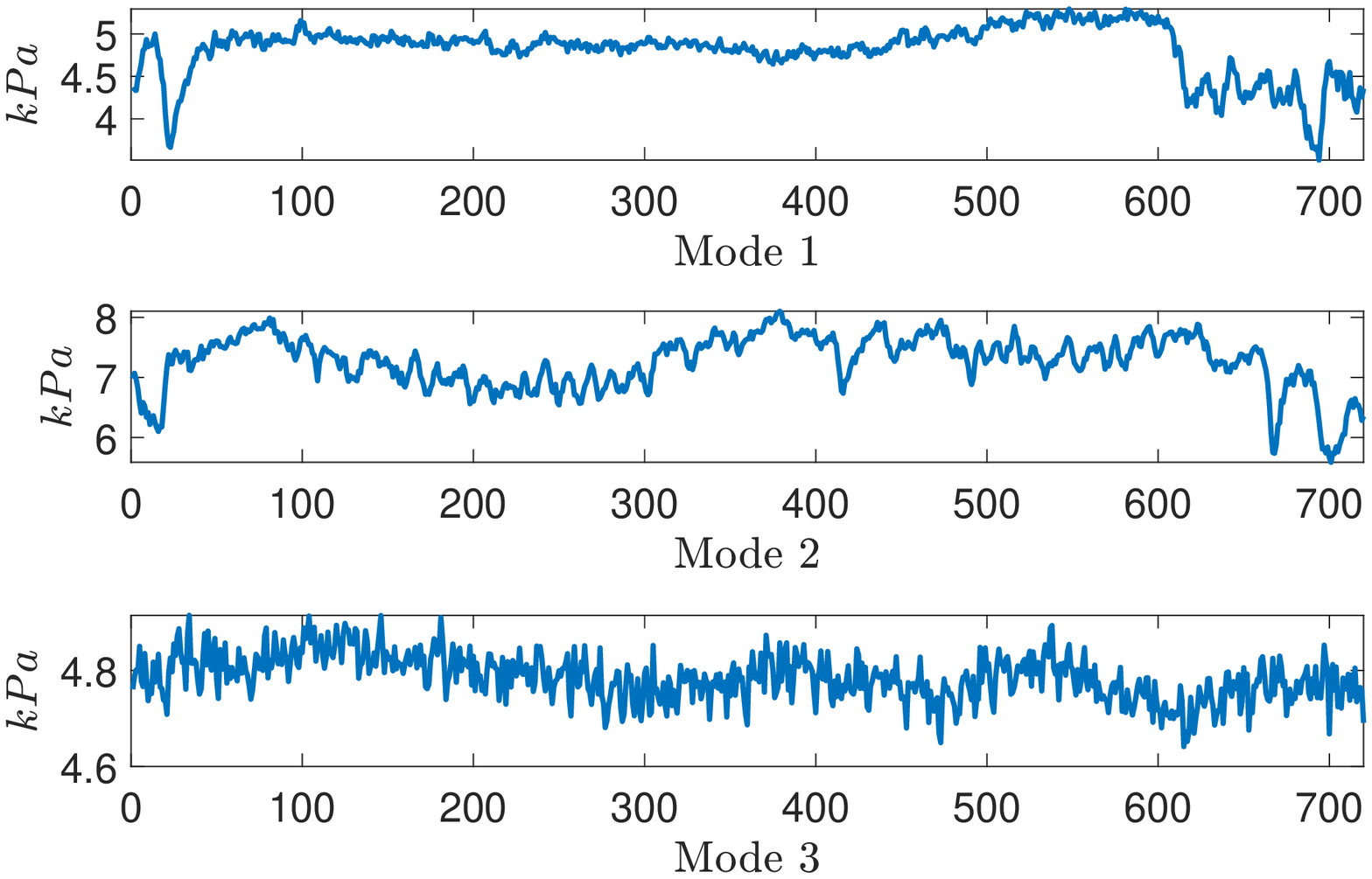}}}
    	\vspace{-1mm}
    	\subfigure{\label{variable-3}}\addtocounter{subfigure}{-2}
    	\subfigure
       	{\subfigure[Pressure of air powder mixture ]{\includegraphics[width=0.235\textwidth]{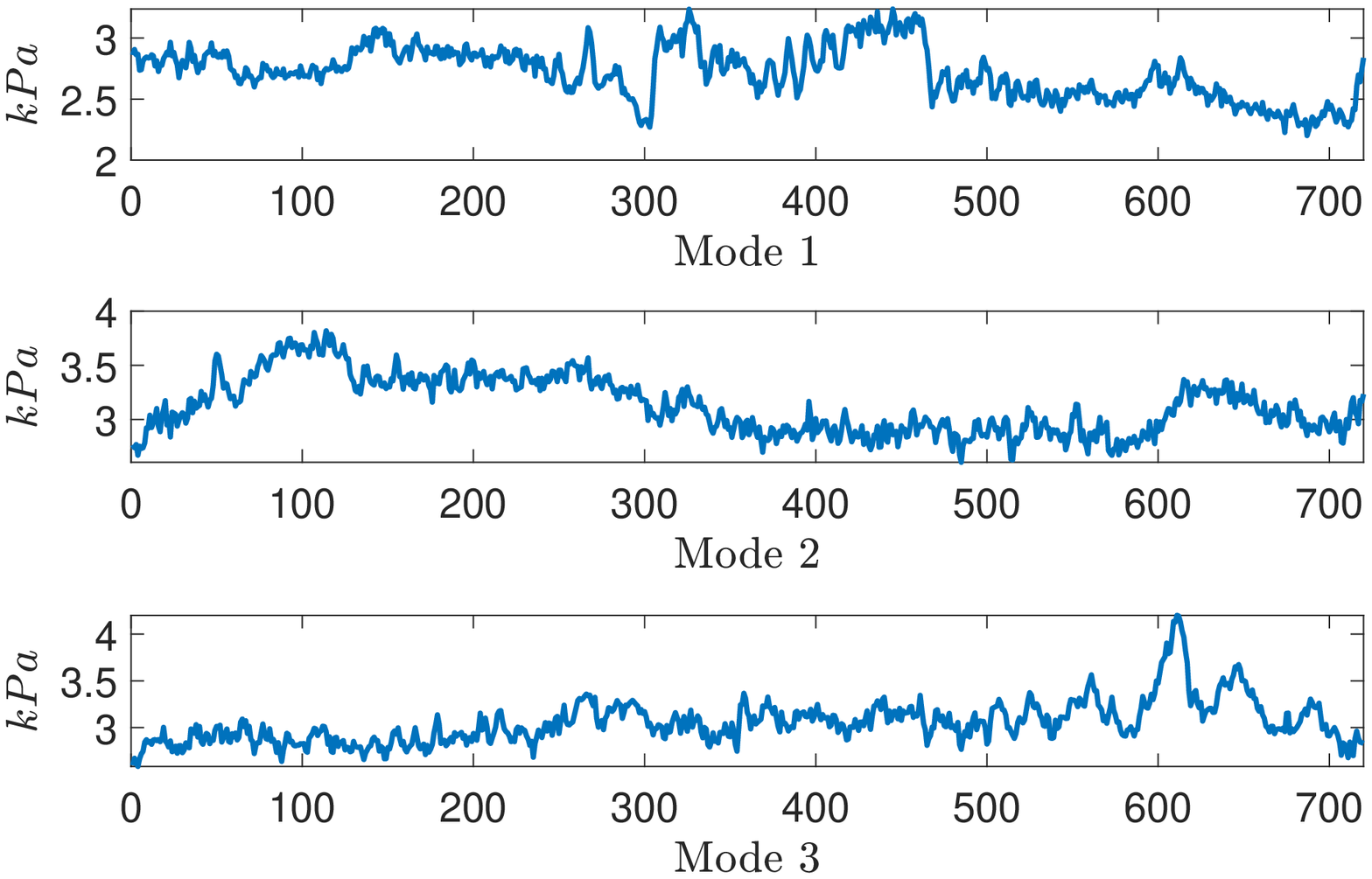}}}
    	\subfigure{\label{variable-4}}\addtocounter{subfigure}{-2}	
    	\subfigure
    	 {\subfigure[Rotating speed of rotary separator]{\includegraphics[width=0.235\textwidth]{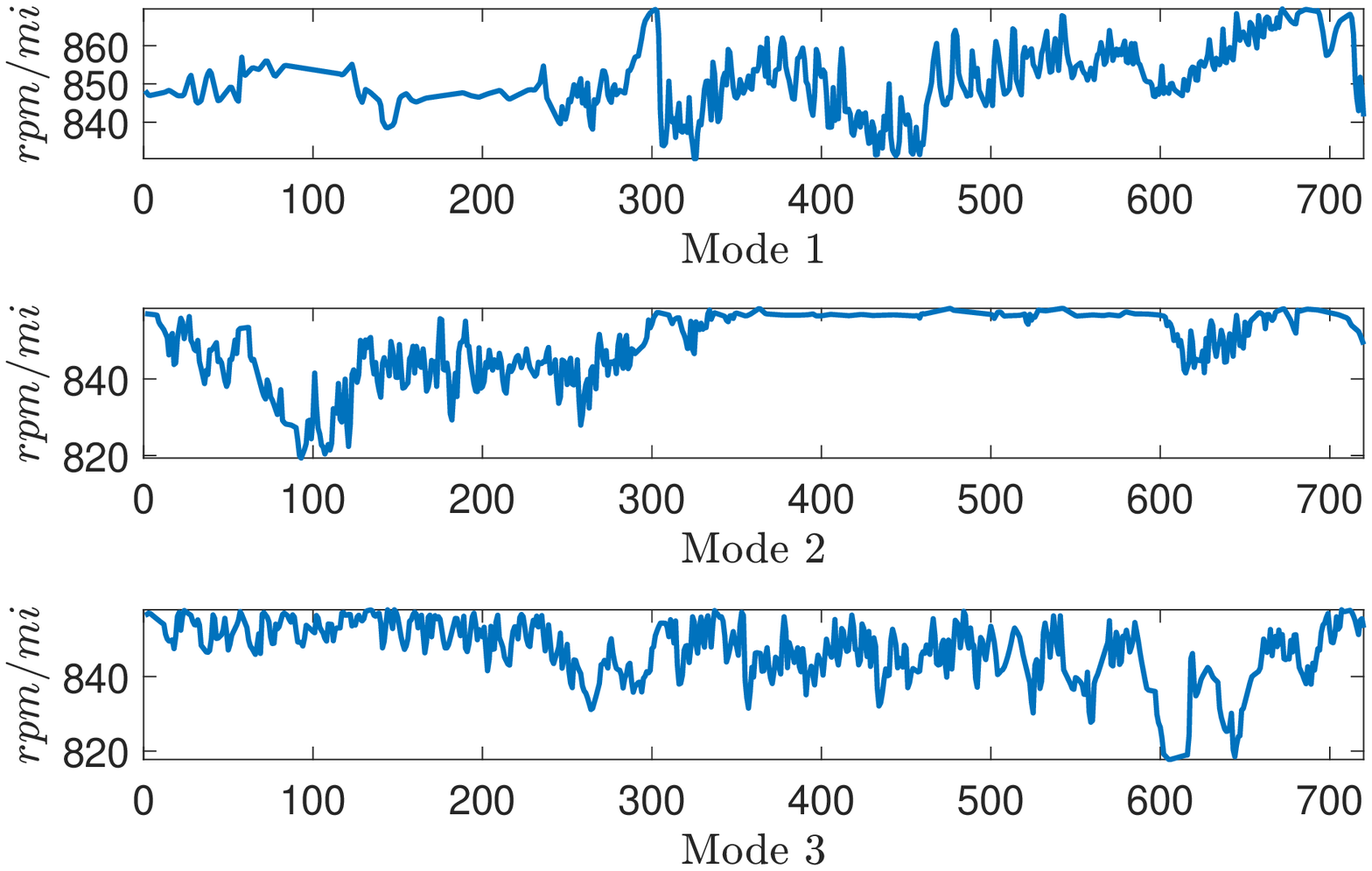}}}
    \caption{Some variables of three modes} \label{variables_figure}
    \end{figure*}

 We select two typical cases to illustrate the effectiveness of PCA-EWC, namely, abnormality from outlet temperature (Fault 4) and rotary separator (Fault 5). Detailed information is listed in Table \ref{Table_informationZS}. The sample interval is 1 minute. Note that the numbers of training samples and testing samples are abbreviated as NoTrS and  NoTeS, respectively.
The training data 1 and 2 come from modes ${\mathcal M}_1$ and ${\mathcal M}_2$, respectively. Testing data $i$,  are collected from mode ${\mathcal M}_i$, $i=1,2,3$. Three modes occurred successively and have different mean values as well as standard deviations.
Partial variables are exhibited in Figure \ref{variables_figure}. We can find intuitively that data from three modes have a certain degree of similarity, especially the modes ${\mathcal M}_1$ and ${\mathcal M}_3$.  Four situations are designed according to Table \ref{Table-simulation}.  Another six situations are designed  as comparative experiments in Table \ref{Table2-comparative}.

Here we discuss the mode identification manners for this case study. Literature  \cite{zhao2021condition} presented a novel mode identification technique, where the coal feed rate  was selected as the condition indicator and then the process was decomposed into several stationary modes automatically. It is effective for the pulverizing system when the type of coal remains the same. For different types of coal, the key variables have different optimal values even though they share the same coal feed rate, such as outlet temperature and the rotary separator speed. In this case study, the mode is identified by the statistical characteristics of partial variables and prior knowledge. Specifically, the  use coal plan is accessible in advance in practical power plants. The set points of outlet temperature and rotary separator speed would change dramatically when the coal switches, which means that the mean values and variances change. Therefore, the appearance of new mode is detected. Besides,  the load is stationary in each mode.

\begin{figure*}[!htp]
	\centering
	\subfigure{\label{fault4-1}}\addtocounter{subfigure}{-2}
	\subfigure
	{\subfigure[Situation 1]{\includegraphics[width=0.235\textwidth]{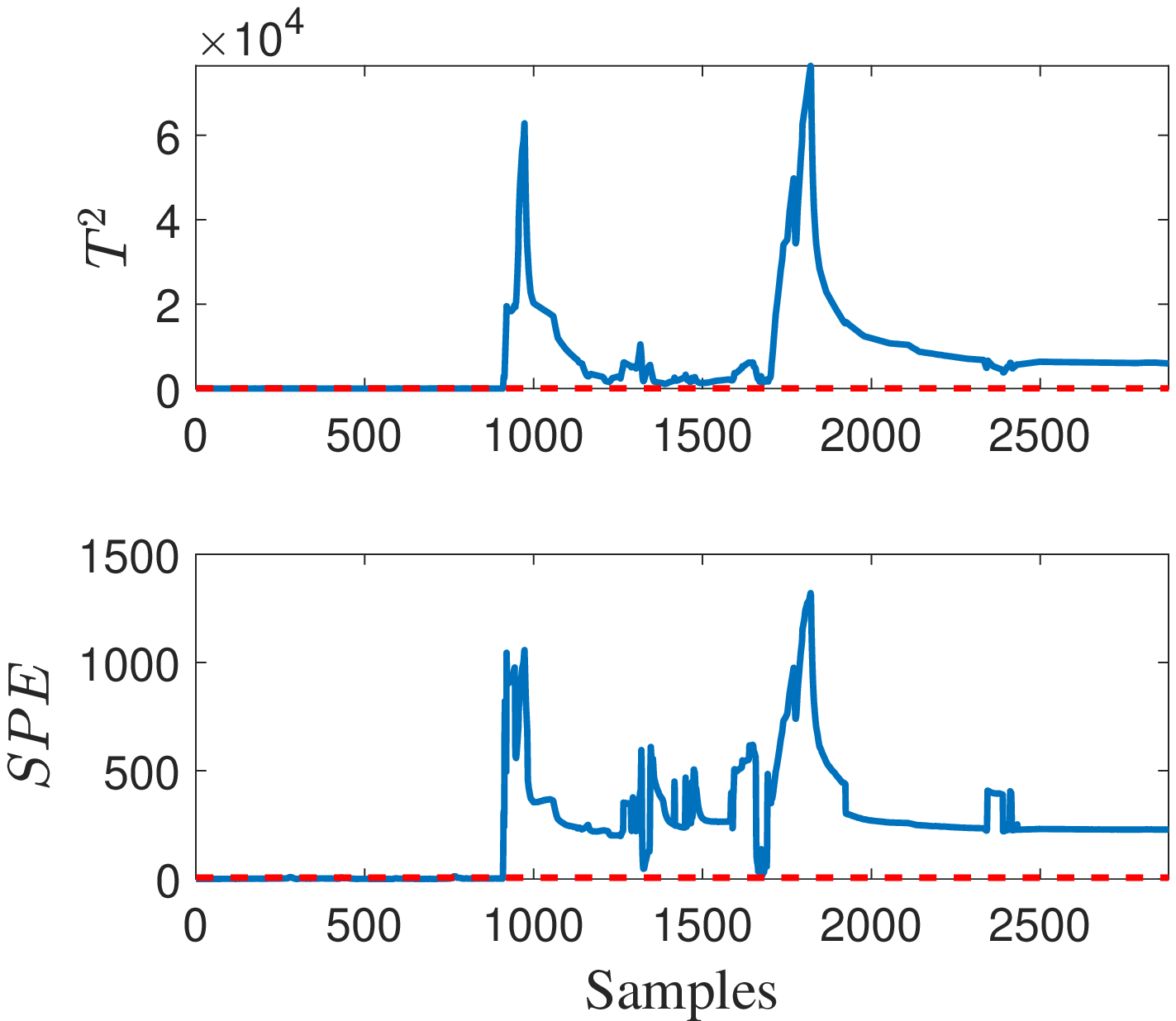}}}
	\hspace{-1mm}
	\vspace{-2mm}
	\subfigure{\label{fault4-2}}\addtocounter{subfigure}{-2}
	\subfigure
	{\subfigure[Situation 2]{\includegraphics[width=0.235\textwidth]{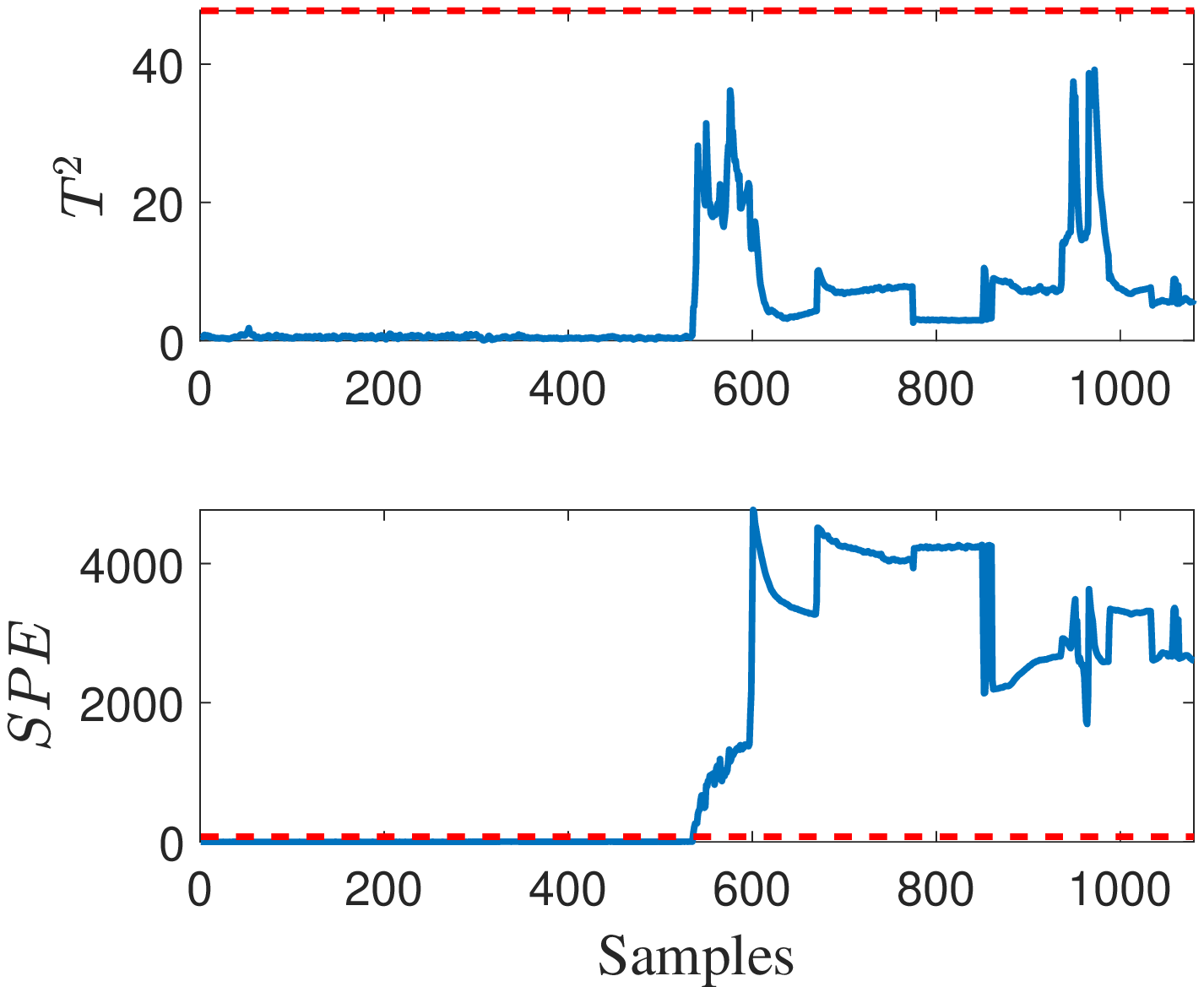}}}
	\vspace{-2mm}
	\hspace{-1mm}
	\subfigure{\label{fault4-3}}\addtocounter{subfigure}{-2}
	\subfigure
	{\subfigure[Situation 3]{\includegraphics[width=0.235\textwidth]{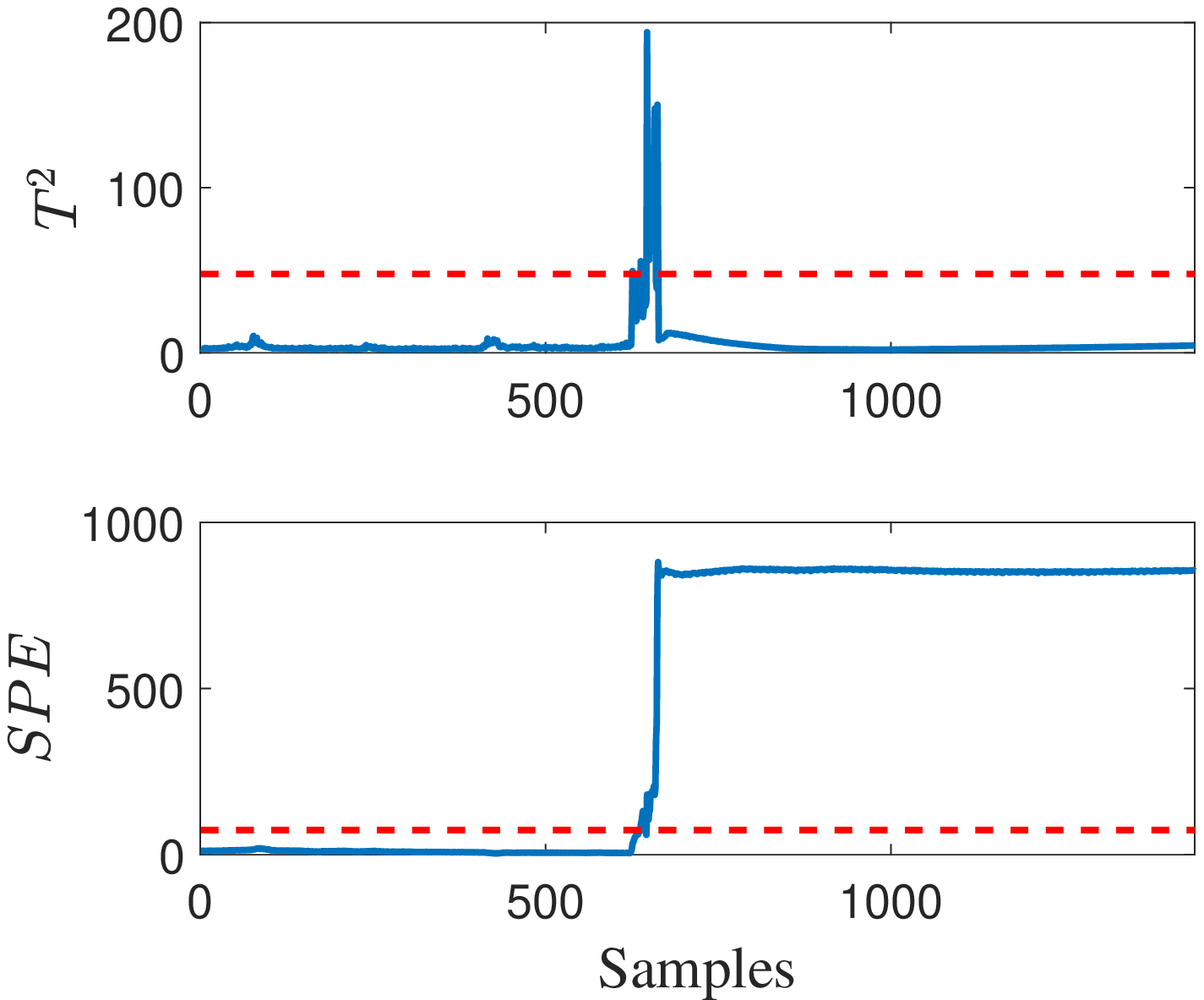}}}
	\subfigure{\label{fault4-4}}\addtocounter{subfigure}{-2}
	\hspace{-1mm}
	\vspace{-2mm}
	\subfigure
	{\subfigure[Situation 4]{\includegraphics[width=0.235\textwidth]{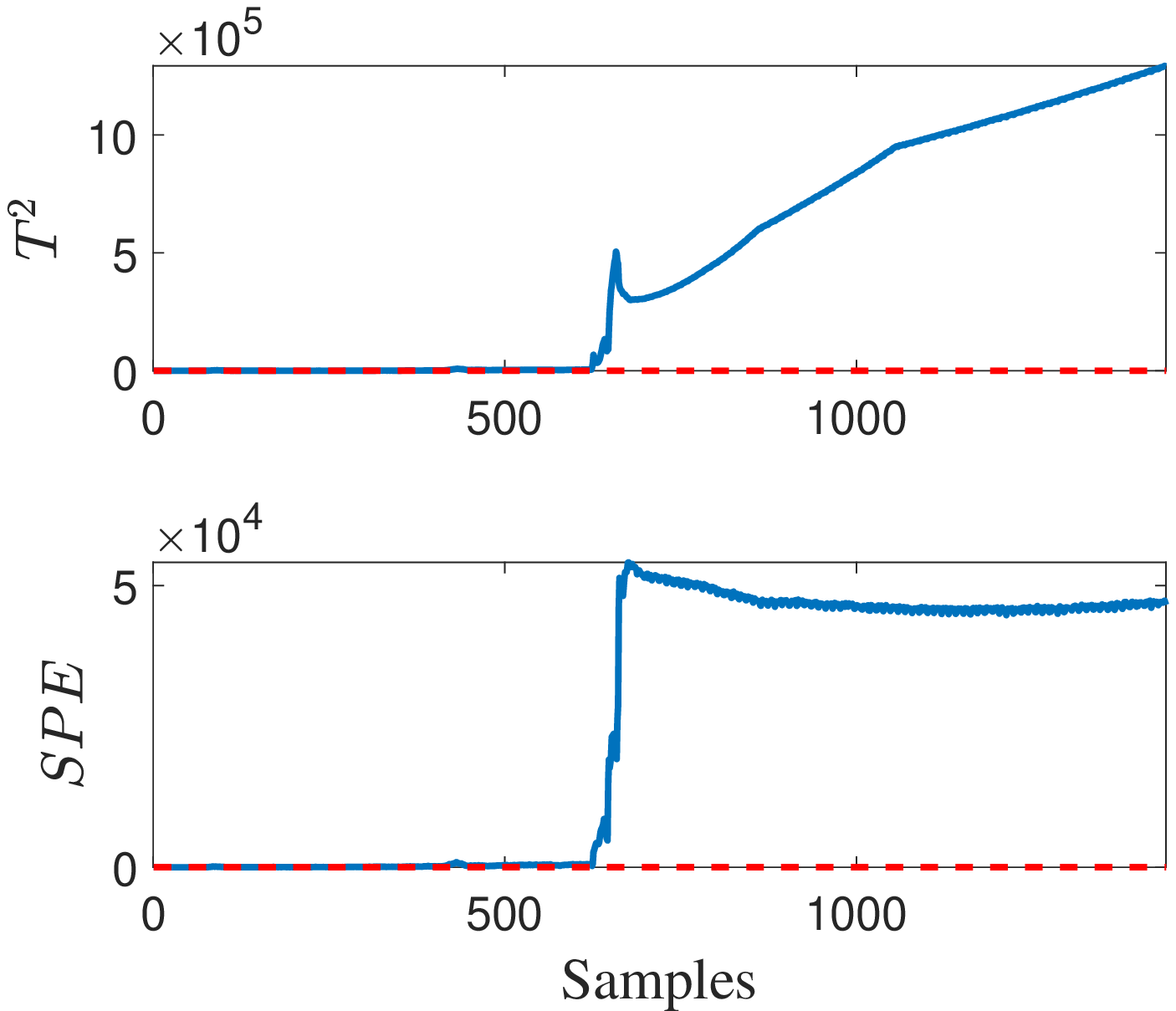}}}
	\subfigure{\label{fault4-5}}\addtocounter{subfigure}{-2}
	\subfigure
	{\subfigure[Situation 5]{\includegraphics[width=0.235\textwidth]{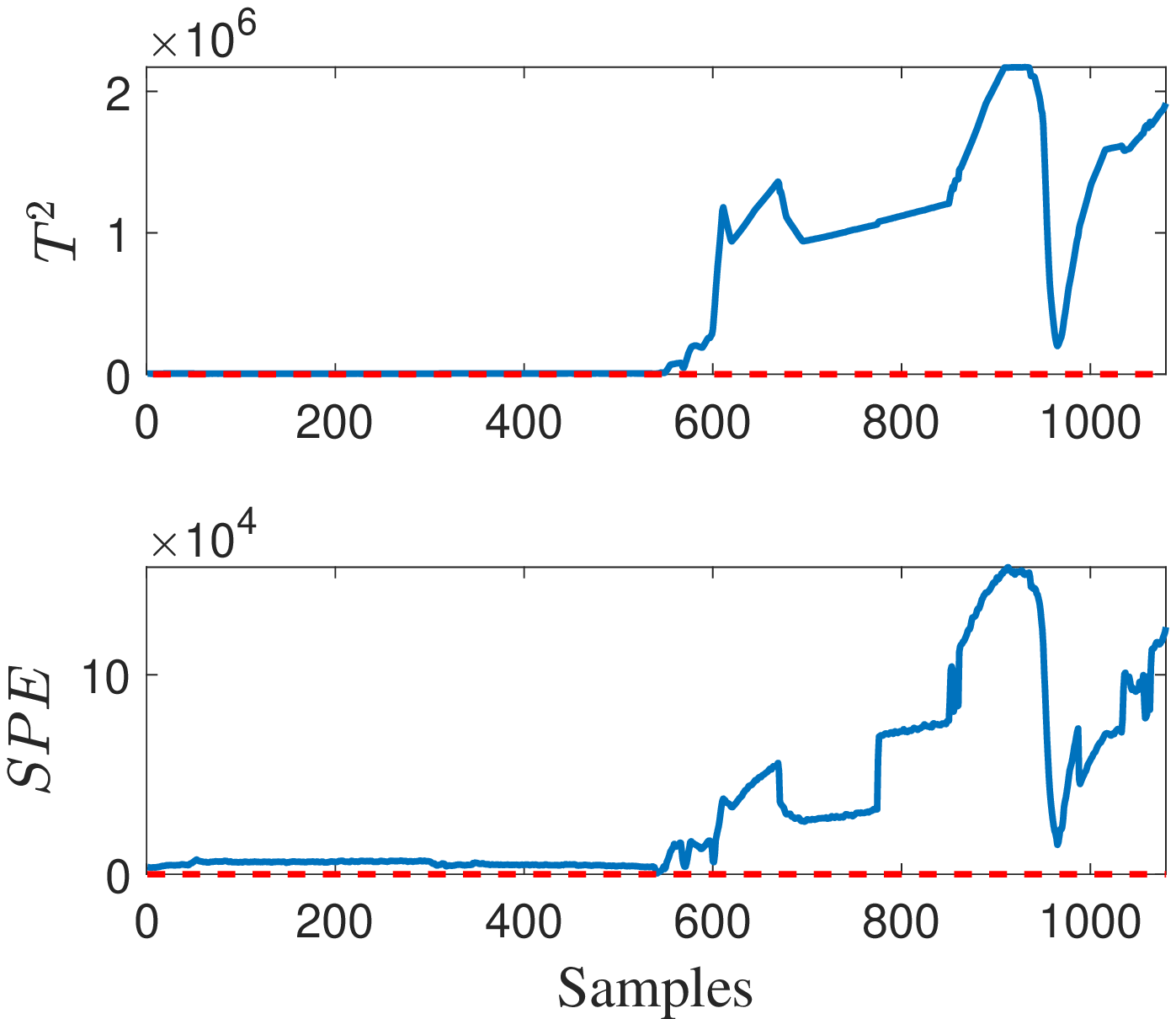}}}
	\subfigure{\label{fault4-6}}\addtocounter{subfigure}{-2}
	\hspace{-1mm}
	\vspace{-2mm}
	\subfigure
	{\subfigure[Situation 6]{\includegraphics[width=0.231\textwidth]{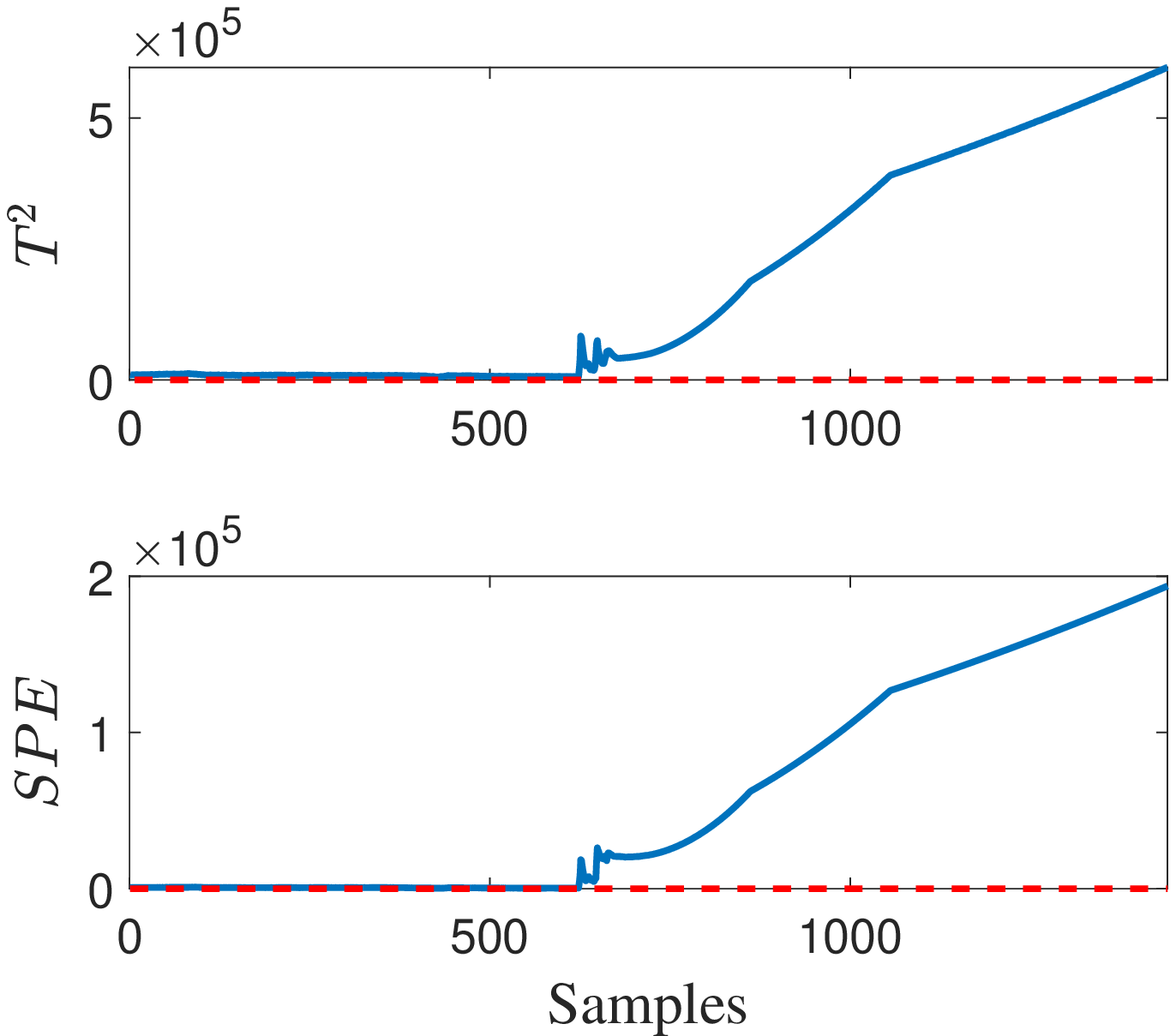}}}
	\subfigure{\label{fault4-7}}\addtocounter{subfigure}{-2}
	\subfigure
	{\subfigure[Situation 7]{\includegraphics[width=0.235\textwidth]{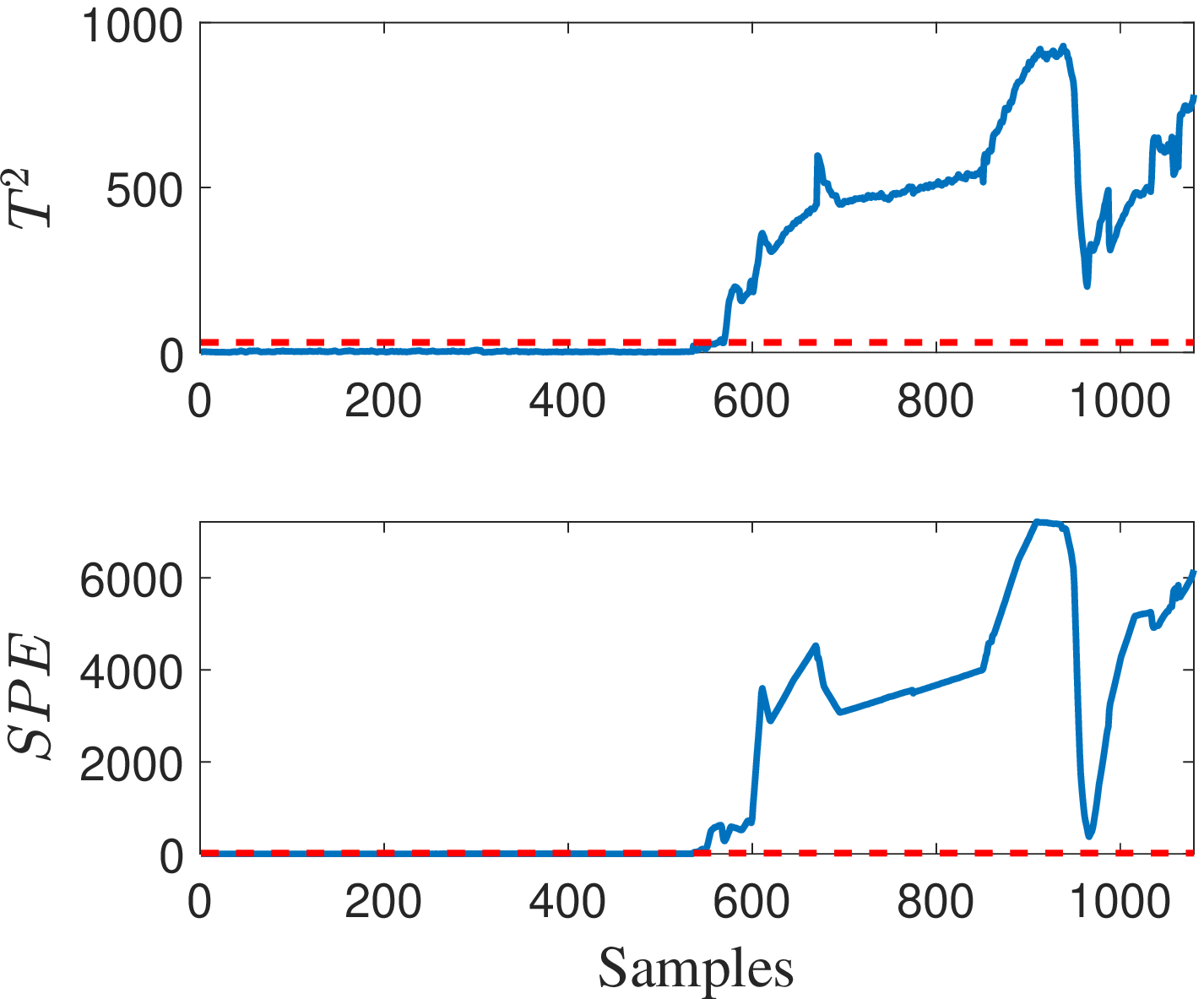}}}
	\hspace{-1mm}
	\vspace{-2mm}
	\subfigure{\label{fault4-8}}\addtocounter{subfigure}{-2}
	\subfigure
	{\subfigure[Situation 8]{\includegraphics[width=0.235\textwidth]{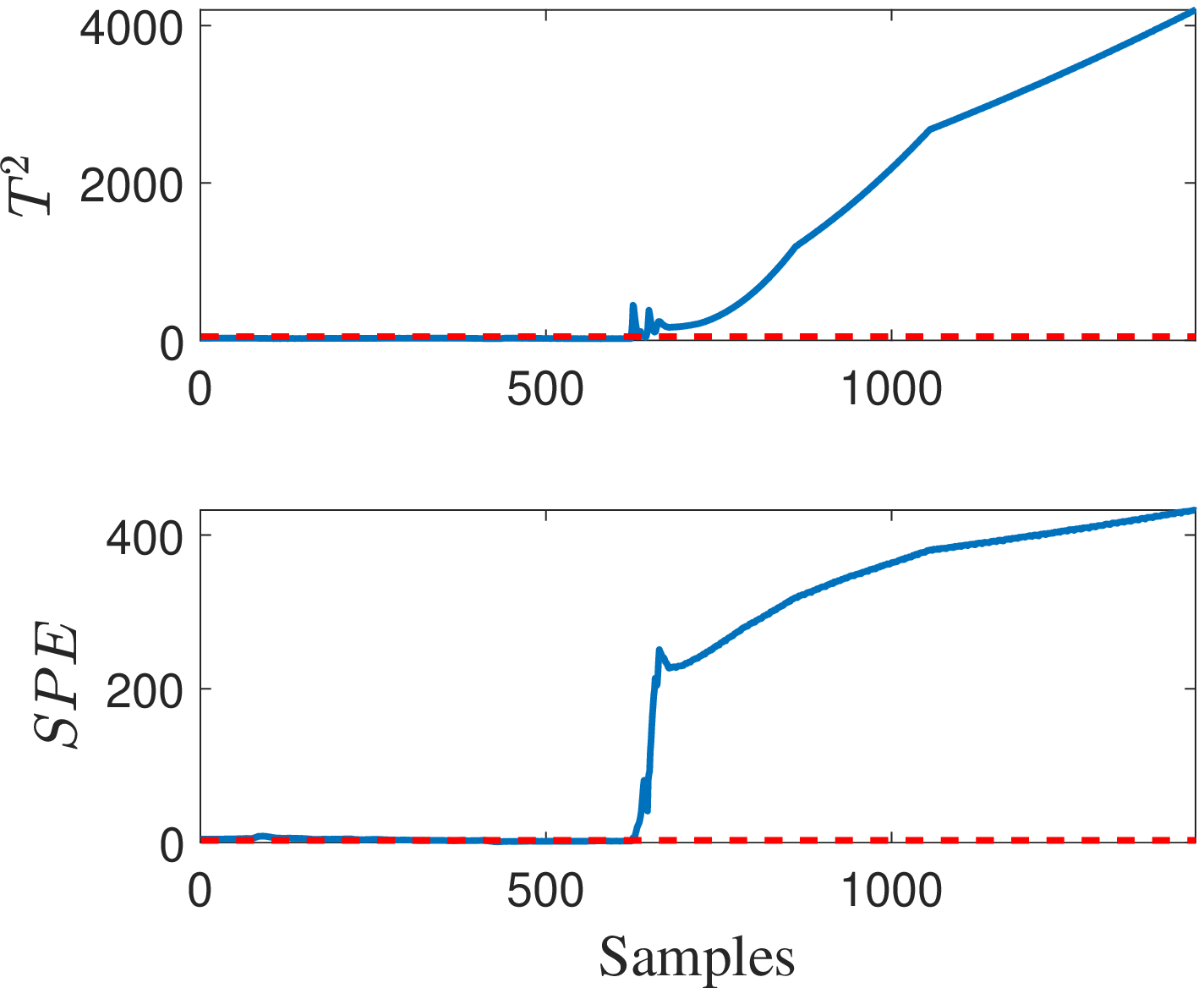}}}
	\hspace{-1mm}
	\vspace{-2mm}
		\subfigure
	{\subfigure[Situation 9]{\includegraphics[width=0.235\textwidth]{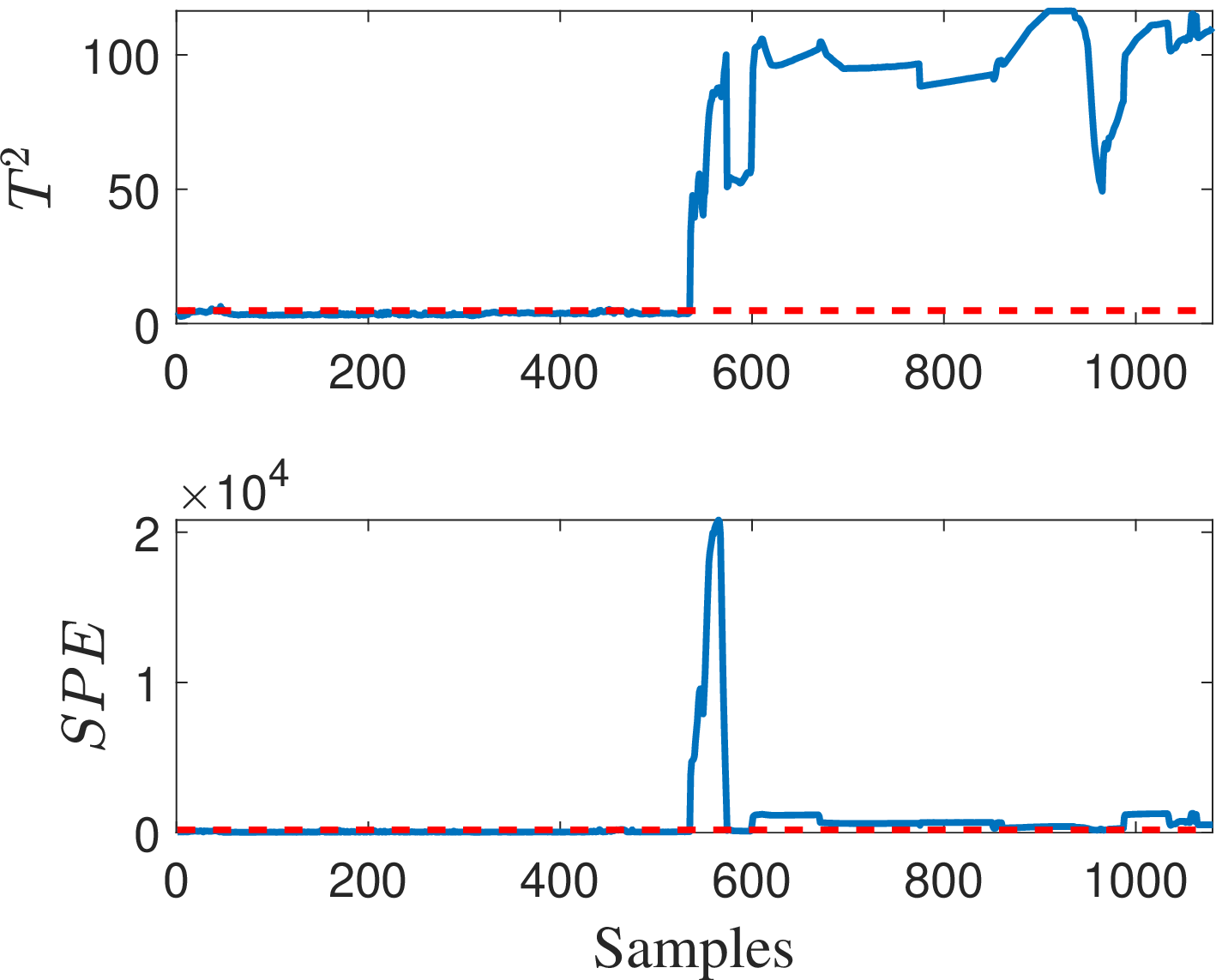}}}
	\subfigure{\label{fault4-9}}\addtocounter{subfigure}{-2}
	\hspace{-1mm}
	\vspace{-2mm}
	\subfigure
	{\subfigure[Situation 10]{\includegraphics[width=0.235\textwidth]{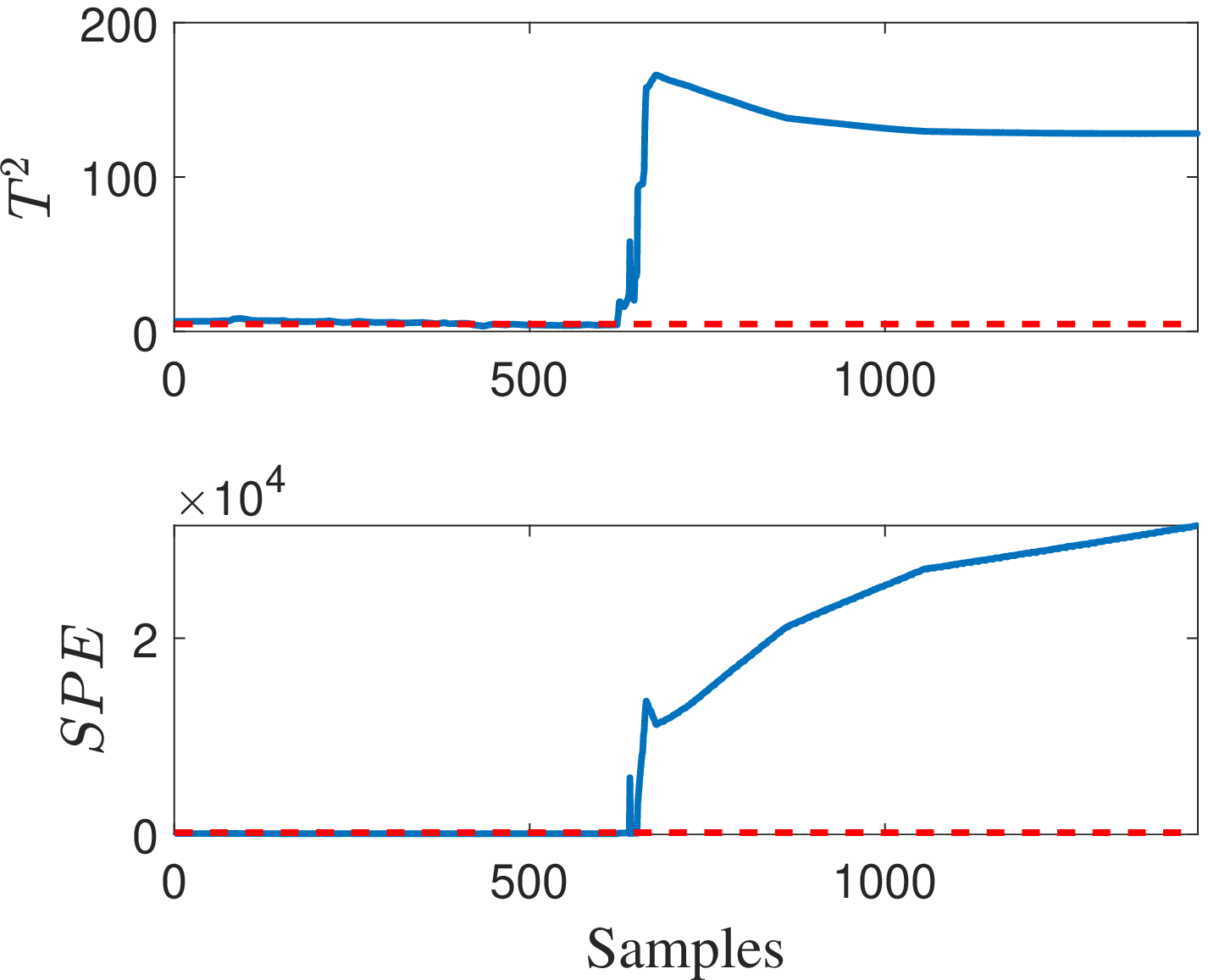}}}
	\caption{Monitoring charts of Fault 4} \label{case4}
\end{figure*}

\begin{table}[!htp]
	\begin{center}
		\caption{Fault detection results of the coal pulverizing system}\label{Table_simulationresultszs}
		\footnotesize
   \begin{tabular}{c c c c c c c c c c c } 
				\hline
		\multirow{1}{*}{Fault type} &\multicolumn{3}{c}{Fault 4}  & \multicolumn{3}{c}{Fault 5}     \\
			\cmidrule(r){1-1}    \cmidrule(r){2-4}  \cmidrule(r){5-7}
		Indexes   &\multicolumn{1}{c} {FDR}    & {FAR}   & \multicolumn{1}{c}{DD}   &\multicolumn{1}{c} {FDR}    &FAR   & \multicolumn{1}{c}{DD}    \\		
			\hline
			Situation 1  & 99.95  &3.19  &1       &100  &0 &0 \\
			Situation 2  & 99.45  & 0   & 3       & 100  & 3.70  & 0 \\
			Situation 3  & 98.40  & 0  & 0        & 88.60  & 0 &  7\\
			Situation 4  & 100    & 91.04 & -     & 100   & 23.32  & - \\
			Situation 5  & 99.45  & 0  &  3       & 100  &5.98  & 0 \\
			Situation 6  & 89.98  & 66.35  & -    &95.63  & 61.54 & - \\
			Situation 7  & 100  & 100  & -        & 100  & 100 & -  \\
			Situation 8  &   100  & 100  & -      & 100  & 100 & -  \\
			Situation 9  & 99.45   & 4.14  & 3     & 100   &17.09  &0  \\
			Situation 10 & 100  &  66.88 &  0      & 100  & 100 & - \\			
			\hline
		\end{tabular}
	\end{center}
\end{table}

The simulation results are summarized in Table \ref{Table_simulationresultszs}. We take the Fault 4 as an instance to interpret the monitoring results specifically.
The monitoring charts of Fault 4 are presented in Figure \ref{case4}. PCA can detect the fault of the mode ${\mathcal M}_1$ and the FDR is $99.95\%$. The detection delay is 1 minute and acceptable. Then, PCA-EWC can also monitor the mode ${\mathcal M}_2$ accurately and the FAR is $0$. Besides, the Model B performs well on the mode ${\mathcal M}_3$ and the FDR is $100\%$, as illustrated in Figure \ref{fault4-3}. It illustrates the continual learning ability of PCA-EWC for monitoring successive modes.
In other words, after training the new mode based on PCA-EWC, we can still acquire excellent performance of the modes, which are similar to previous trained modes. In Figure \ref{fault4-4}, the FAR is more than $90\%$ and the training Model C fails to detect the fault. It is meaningless to mention FDR and detection delay here. Furthermore, it verifies that PCA completely forgets the information of previous modes and the previous learned knowledge is over-written by new information, thus leading to an abrupt performance decrease for monitoring previous modes.  In Figures \ref{fault4-5} and \ref{fault4-6}, RPCA is not capable of tracking the successive modes and detecting faults accurately. GMMs can provide excellent performance for the trained mode and the FDR is $99.45\%$, as illustrated in Figure \ref{fault4-7}. However, GMMs fail to monitor the similar mode that is not trained before and the FAR is $66.35\%$ in Figure \ref{fault4-8}. For Situations 9 and 10, similar to numerical case study, IMPPCA enables to detect the fault in the previous trained mode accurately but fails to monitor the similar mode.
The above-mentioned analysis of Fault 4 applies equally to Fault 5. Although the FAR of Situation 4 is much lower than that of Fault 4, it is over $20\%$ and not acceptable.

\begin{figure}[!htbp]
	\centering
	\subfigure{\label{rawdata}}\addtocounter{subfigure}{-2}
	\subfigure
	{\subfigure[Similarity of original data]{\includegraphics[width=0.232\textwidth]{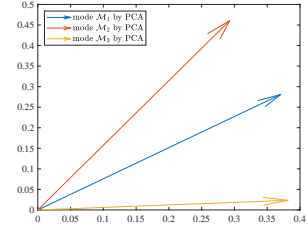}}}
	\subfigure{\label{immediateresult}}\addtocounter{subfigure}{-2}
	\subfigure
	{\subfigure[Effect of PCA-EWC]{\includegraphics[width=0.232\textwidth]{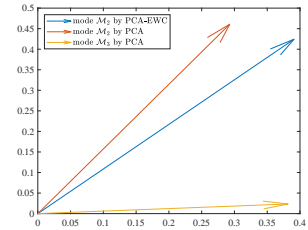}}}
	\centering
	\caption{Illustrations of superiorities of PCA-EWC} \label{immediate_variable5}  
\end{figure}

Here we present the simulation results of PCA-EWC to verify the superiorities of PCA-EWC further.
To represent the immediate results intuitively, the first two components of projection vectors are shown in Figure \ref{immediate_variable5}. The normal data from Fault 4 are employed to train the PCA and PCA-EWC models.
The cosine similarity measure is adopted to evaluate the similarity.
Modes ${\mathcal{M}_1}$ and ${\mathcal{M}_3}$ have a certain degree of similarity while ${\mathcal{M}_2}$ and ${\mathcal{M}_3}$ are different, as described in Figure \ref{rawdata}.
If the single monitoring model is established based on the traditional approaches, the features of mode ${\mathcal{M}_1}$ are over-written. When the mode ${\mathcal{M}_3}$ arrives, the current model based on data from mode ${\mathcal{M}_2}$ fails to provide the excellent performance. However, when PCA-EWC is adopted to preserve the significant features of mode  ${\mathcal{M}_1}$, the projection vector is similar to that of mode ${\mathcal{M}_3}$ in Figure \ref{immediateresult}, thus delivering the outstanding monitoring performance.

 According to the comparison among 10 situations, PCA-EWC can preserve significant information of previous modes and it is enough to monitor the new similar modes precisely.   However, traditional PCA forgets most information of previous monitoring tasks when a PCA-based model is built for the current mode, and thus the performance is decreased catastrophically.  RPCA, GMMs and IMPPCA fail to monitor the successive modes and the monitoring mode needs to be retrained for GMMs and IMPPCA.

\section{Conclusions}\label{Sec:5}
This paper provides a novel framework PCA-EWC with continual learning ability for monitoring multimode processes.
The proposed algorithm adopts EWC to tackle the catastrophic forgetting of significant features of previous modes if PCA-based monitoring models are used, and the geometric illustration is depicted to understand the core thoroughly. The designed objective function of PCA-EWC is nonconvex and reformulated as DC programming, where the optimal solution is calculated thereafter. The spirit of PCA-EWC has been extended to more general multimode processes and the detailed implementing procedure  has been presented.
Moreover, it has been interpreted that PCA-EWC is reformulated as PCA by specific parameter setting and slightly complicated than PCA. Besides, PCA-EWC preserves partial information of previous modes when modified PCA is utilized for the current mode. However, traditional PCA-based models can't remember the significant features of previous modes, and thus resulting in the abrupt decrease of performance if single local monitoring model is used. The potential limitation is the existence of similarity among modes and the solution is provided.
Compared with RPCA, GMMs and IMPPCA, the effectiveness of PCA-EWC has been illustrated by a numerical case and a practical industrial system.

In future, we will resolve the problem that the number of principal components remains the same,  multimode nonstationary process monitoring with continual learning ability and the transition process between two successive modes would be investigated also.  Besides, to monitor the process with frequent mode changes accurately,  simultaneous static and dynamic analysis would be considered.

\appendix
\section{Details of Laplace approximation}\label{sec:2.2}
Assume that model for mode $ {\mathcal{M}_1} $ has been already built when data $\boldsymbol X_2$ are collected. The optimal parameter for mode $ {\mathcal{M}_1} $ is
\begin{equation}
	{\theta _{{\mathcal{M}_1}}^*} = \arg\min_{\theta} \; \{-\log p(\theta|\boldsymbol X_1)\}
\end{equation}
Obviously,  $ {\partial \log p(\theta|\boldsymbol X_1)}/{\partial \theta} =0 $ at $ {\theta _{{\mathcal{M}_1}}^*} $. Based on the second order Taylor series around  $ {\theta _{{\mathcal{M}_1}}^*} $, $ -\log p(\theta|\boldsymbol X_1) $  can be approximated as
\begin{equation}\label{secondtaylor}
	-\log p(\theta|\boldsymbol X_1) \approx \frac{1}{2}(\theta-{\theta _{{\mathcal{M}_1}}^*})^T \boldsymbol H({\theta _{{\mathcal{M}_1}}^*})(\theta-{\theta _{{\mathcal{M}_1}}^*})+ constant
\end{equation}
where $\boldsymbol H({\theta _{{\mathcal{M}_1}}^*}) $ is the Hessian matrix  of $ -\log p(\theta|\boldsymbol X_1) $ with respect to $\theta$ at $ {\theta _{{\mathcal{M}_1}}^*} $ \cite{huszar2017on}. Since ${\theta _{{\mathcal{M}_1}}^*}$ is a local minimum, $ \boldsymbol H({\theta _{{\mathcal{M}_1}}^*}) $  is  approximated by
\begin{equation}\label{hessianmatrix}
	\boldsymbol H({\theta _{{\mathcal{M}_1}}^*}) \approx N_1  \boldsymbol F({\theta _{{\mathcal{M}_1}}^*})+ \boldsymbol H_{prior}({\theta _{{\mathcal{M}_1}}^*})
\end{equation}
where $N_1$ is the number of  data $\boldsymbol X_1$, $\boldsymbol F({\theta _{{\mathcal{M}_1}}^*}) $ is the Fisher
information matrix for mode $ {\mathcal{M}_1} $,  $\boldsymbol H_{prior} $ is the Hessian matrix of $ -\log p(\theta) $ and $ p(\theta) $ is the prior of parameters \cite{huszar2017on}.  Here we assume that the prior is an isometric Gaussian prior, and $\boldsymbol H_{prior}({\theta _{{\mathcal{M}_1}}^*}) = \lambda_{prior} \boldsymbol I $ is adopted. According to (\ref{secondtaylor}-\ref{hessianmatrix}), the Laplace approximation of (\ref{poster2}) is reformulated as
\begin{equation}\label{laplaceapproximation}
	\begin{aligned}
		\log \,p\left( {\theta |\boldsymbol X} \right) \approx& \log \,p\left( {\boldsymbol X_2|\theta } \right) -\frac{1}{2}(\theta-{\theta _{{\mathcal{M}_1}}^*})^T\\
		&  (N_1 \boldsymbol F_{{\mathcal{M}_1}}+\lambda_{prior} \boldsymbol I)(\theta-{\theta _{{\mathcal{M}_1}}^*})+ constant
	\end{aligned}
\end{equation}
However, the sample size $N_1$ has significant influence on quality of approximation \cite{huszar2017on}. A hyper-parameter $\lambda_{\mathcal{M}_1}$ is introduced to control the approximation better \cite{huszar2017on}, namely,
\begin{equation}\label{laplaceapproximation1}
	\begin{aligned}
		\log \,p\left( {\theta |\boldsymbol X} \right) \approx& \log \,p\left( {\boldsymbol X_2|\theta } \right) -\frac{1}{2}(\theta-{\theta _{{\mathcal{M}_1}}^*})^T\\
		&  (\lambda_{\mathcal{M}_1} \boldsymbol F_{{\mathcal{M}_1}}+\lambda_{prior} \boldsymbol I)(\theta-{\theta _{{\mathcal{M}_1}}^*})- constant
	\end{aligned}
\end{equation}

\section{Recursive Laplace approximation}\label{RecursiveLA}
Recursive Laplace approximation \cite{huszar2017on} is employed to approximate the objective function  (\ref{thirdmode}).

The posterior probability
$ \log \,p\left( {\theta |\boldsymbol X_1,\boldsymbol X_2} \right) $
is approximated by (\ref{ewc_final}).
Similar to Appendix \ref{sec:2.2}, Taylor series approximation is applied around $\theta _{\mathcal{M}_2}^*$ and the first order deviation is zero. The Hessian matrix of $ -\log \,p\left( {\boldsymbol X_2|\theta} \right) $  can be approximated and replaced by $\lambda_{\mathcal{M}_2} \boldsymbol F_{\mathcal{M}_2}$, where $ \boldsymbol F_{\mathcal{M}_2} $ is the Fisher information matrix of data $\boldsymbol X_2$ \cite{huszar2017on}. Besides, the second derivative of quadratic penalty is ${\boldsymbol \Omega}_{\mathcal{M}_1}$. Thus, (\ref{thirdmode}) is approximated as:
\begin{equation}\label{ewc_final_thirdmode}
	\begin{aligned}
		\log \,p\left( {\theta |\boldsymbol X} \right) \approx& \log \,p\left( {\boldsymbol X_3|\theta } \right) -(\theta-{\theta _{{\mathcal{M}_2}}^*})^T\\
		&  {\boldsymbol \Omega}_{\mathcal{M}_2} (\theta-{\theta _{{\mathcal{M}_2}}^*})+constant
	\end{aligned}
\end{equation}
where
\begin{equation}
	\begin{aligned}
		{\boldsymbol \Omega}_{\mathcal{M}_2} 
		= {\boldsymbol \Omega}_{\mathcal{M}_1}+\frac{1}{2}\lambda_{\mathcal{M}_2} \boldsymbol F_{{\mathcal{M}_2}}
	\end{aligned}
\end{equation}

\section*{Acknowledgements}

This work was supported by National Natural Science Foundation of China [grant numbers 62033008,  61873143]

\bibliographystyle{model1-num-names}
\bibliography{my_references}



%

\end{document}